\documentclass[11pt]{article}

\usepackage[final]{acl}

\usepackage{times}
\usepackage{latexsym}

\usepackage{enumitem}

\usepackage[T1]{fontenc}

\usepackage[utf8]{inputenc}

\usepackage{microtype}

\usepackage{inconsolata}

\usepackage{graphicx}
\usepackage{booktabs}
\usepackage{multirow}
\usepackage{colortbl}
\usepackage{amsfonts}
\usepackage{subcaption}
\usepackage{tcolorbox}
\definecolor{mydarkpink}{HTML}{c2255c}
\definecolor{mypink}{HTML}{e64980}
\usepackage{amsmath} 
\usepackage{makecell}
\usepackage{adjustbox}

\usepackage[table]{xcolor}
\usepackage{array}
\newcolumntype{G}{>{\columncolor{gray!20}}c}

\newcommand{\ourmethod}[1]{\textsc{FLIP}}

\newcommand{\ignore}[1]{}

\title{Small Reward Models via Backward Inference}

\newcommand{\uw}{$^{\heartsuit}$}
\newcommand{\aitwo}{$^{\spadesuit}$}
\author{Yike Wang\uw \ \ \ Faeze Brahman\aitwo \ \ \ Shangbin Feng\uw \ \ \ Teng Xiao\aitwo \ \ \ \\
\textbf{Hannaneh Hajishirzi}\uw\aitwo \ \ \ \textbf{Yulia Tsvetkov}\uw \ \ \
\\ \uw{}University of Washington \aitwo{}Allen Institute for Artificial Intelligence \\
\texttt{yikewang@cs.washington.edu}
}

\begin{document}
\maketitle
\begin{abstract}
Reward models (RMs) play a central role throughout the language model (LM) pipeline, particularly in non-verifiable domains. 
However, the dominant LLM-as-a-Judge paradigm relies on the strong reasoning capabilities of large models, while alternative approaches require reference responses or explicit rubrics, limiting flexibility and broader accessibility.
In this work, we propose \ourmethod{} (FLipped Inference for Prompt reconstruction), a reference-free and rubric-free reward modeling approach that reformulates reward modeling through backward inference: inferring the instruction that would most plausibly produce a given response
. The similarity between the inferred and the original instructions is then used as the reward signal. 
Evaluations across four domains using 13 small language models 
show that \ourmethod{} outperforms LLM-as-a-Judge baselines by an average of 79.6\%. Moreover, \ourmethod{} substantially improves downstream performance in extrinsic evaluations under test-time scaling via parallel sampling and GRPO training. We further find that \ourmethod{} is particularly effective for longer outputs and robust to common forms of reward hacking. By explicitly exploiting the validation–generation gap, \ourmethod{} enables reliable reward modeling in downscaled regimes where judgment methods fail.
\footnote{Code available at \href{https://github.com/yikee/FLIP}{https://github.com/yikee/FLIP}.}
\end{abstract}

\section{Introduction}


Reward models (RMs) are widely used throughout the language model (LM) pipeline: they play a central role in reinforcement learning, preference optimization, reranking, and automatic evaluation, particularly in non-verifiable domains~\citep{ouyang2022training, rafailov2023direct, brown2024large, snell2024scaling, zhang2024generative, yu2025rewardanything}. 
A dominant approach instantiates RMs via direct judgment, 
a.k.a.~\emph{LLM-as-a-Judge} \citep{zheng2023judging}. In this paradigm, a prompted LM evaluates responses and produces scalar scores, preference labels, or textual critiques that are used as reward signals. 
In practice, the reliability of such judge-based method depends heavily on the superior reasoning capabilities of larger base models \citep{zheng2023judging, gu2024survey}. When applied to small language models (SLMs),\footnote{We define small language models as language models with 8B parameters or fewer.}  performance drops by 41\% according to the RewardBench2 leaderboard \citep{malik2025rewardbench}.
The cost is also amplified by the fact that RMs are invoked repeatedly during optimization and evaluation, often dominating the overall compute footprint~\citep{ouyang2022training, bai2022training}. 
These considerations motivate our goal: \textit{to develop reward modeling methods that remain reliable and effective when downscaled to small general-purpose language models.}

\begin{figure}[t]
    \centering
    \includegraphics[width=\linewidth]{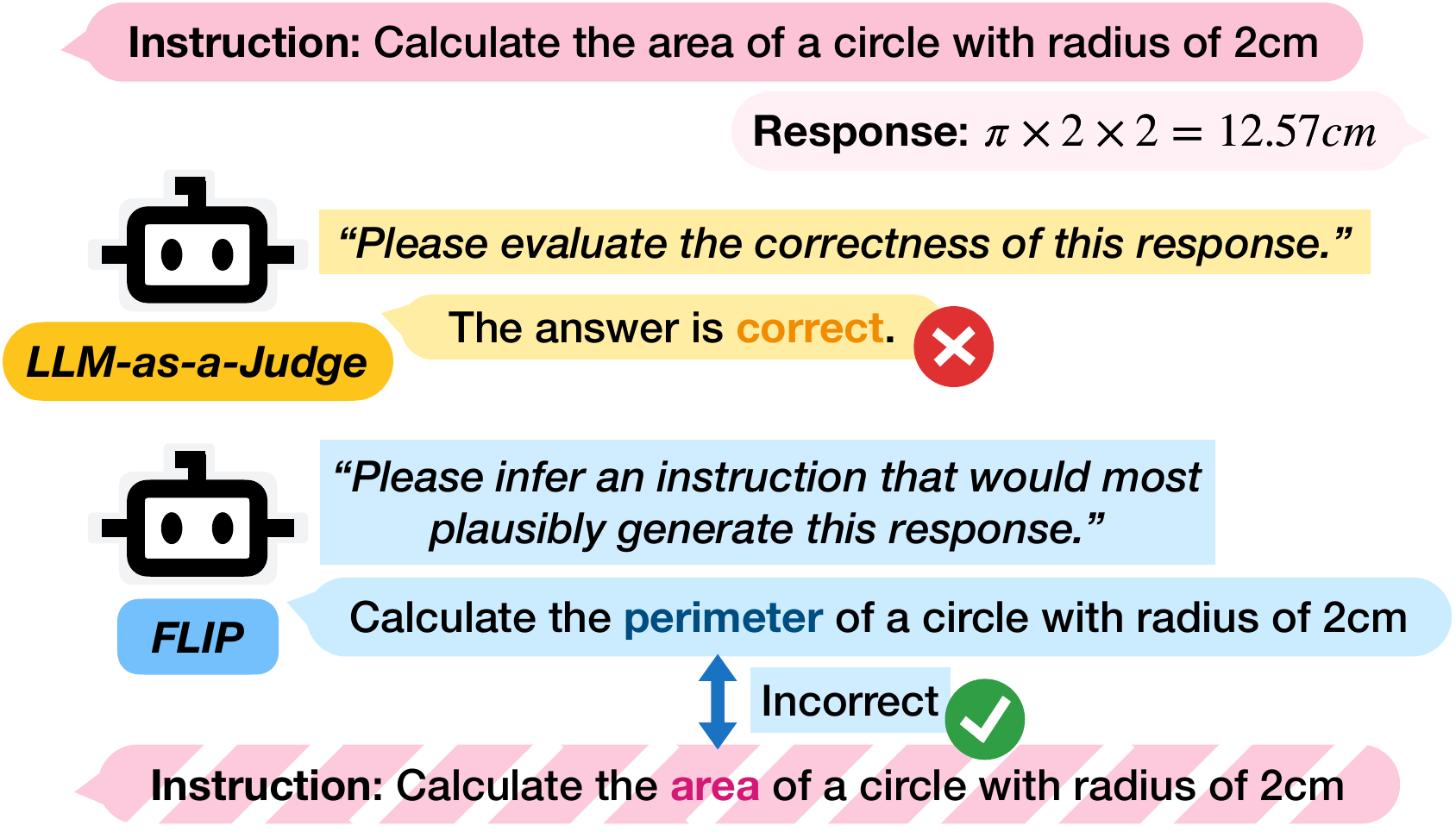}
    \caption{While LM judges can be misled, \ourmethod{} effectively identifies off-topic, instruction-misaligned, and factually incorrect responses via backward inference. 
    }
    \label{fig:teaser}
    \vspace{-4mm}
\end{figure}

In this work, we argue that effective and efficient reward modeling in downscaled regimes requires a reformulation of the judgment paradigm itself. 
Based on the intuition that given a high-quality response with sufficient length and contextual richness, the original query should be able to be inferred from it, we introduce \ourmethod{} (FLipped Inference for Prompt reconstruction), 
a reward modeling approach based on backward inference---\emph{inferring the instruction that would most plausibly produce a given response}. Given a response, a language model is used to infer the instruction under which the response would most likely arise, and \emph{the similarity between the inferred instruction and the original instruction is used as the reward signal} (Figure~\ref{fig:teaser}). 
We explain our method through Bayesian theory~\citep{bernardo1994bayesian}. 

Experimental results on a standard reward modeling benchmark across four domains and 13 SLMs spanning multiple families and sizes show that SLMs with backward inference are strong reward models.
Our approach consistently surpasses judgment baselines by an average of 79.6\%, and substantially outperforms them in extrinsic evaluations under test-time scaling via parallel sampling across four benchmarks, as well as Group Relative Policy Optimization (GRPO) training across five benchmarks. 
\ourmethod{} is well-suited for downscaled regimes as it does not rely on reference responses or rubrics, which are costly and difficult to obtain.
We further find that \ourmethod{} is particularly effective on longer outputs and is robust to common forms of reward hacking
. 
We hypothesize that the effectiveness of \ourmethod{} stems from the validation--generation gap \citep{west2024the, li2023benchmarking}, which is more pronounced in smaller language models: while small models often struggle with judgment, they remain relatively strong in generative inference. 

In summary, 
(1) we empirically examine reward modeling in downscaled settings and demonstrate that judgment becomes unreliable for small models; (2) we propose \ourmethod{}, a generative reformulation of reward modeling that is reference-free and rubric-free and explicitly designed for SLMs; and (3) we show through extensive intrinsic and extrinsic evaluations that \ourmethod{} substantially outperforms judgment baselines. We hope that our findings encourage the community to rethink how judgment is operationalized in reward modeling and 
to further pursue cost-efficient training and inference of language models in downscaled regimes \citep{belcak2025small, goel2025position}.

\section{Reward Modeling}
In this section, we present the formulation of reward modeling in \ref{sec:problem_setting}, our proposed method \ourmethod{} in \ref{sec:our_method}, as well as baseline settings in \ref{sec:variants}. We later showcase the efficacy of our method on reward modeling benchmarks (Section~\ref{sec:intrinsic_eval_setting}), in test-time scaling (Section~\ref{sec:bon_setting}), and in Reinforcement Learning (Section~\ref{sec:grpo_setting}).

\label{sec:method}
\subsection{Problem Setting}
\label{sec:problem_setting}
Let \(x\) denote an instruction and \(y\) a response generated by a language model parametrized by \(\theta\), \[ y \sim p_\theta(y \mid x). \] 
Given 
a language model parameterized by \(\phi\) as the reward model, our goal is to estimate a scalar reward \(r \in \mathbb{R}\) that reflects the quality of the response. We define this reward via a scoring function 
\[
r(x, y) = f_1(\mathrm{LM}_{\phi},x,y),
\]
where \(f_1(\cdot)\) maps \(\mathrm{LM}_{\phi}\) and the pair \((x,y)\) to a real-valued score capturing attributes such as correctness, helpfulness, and overall response quality. 

In some cases, we are given a set of candidate responses ${y_1, y_2, \ldots, y_n}$ 
and our goal is to estimate a ranking (i.e., a preference ordering) over these responses using the reward model \(\mathrm{LM}_{\phi}\).
Specifically, we seek a permutation \(\pi\) such that
\[
y_{\pi(1)} \succ y_{\pi(2)} \succ \cdots \succ y_{\pi(n)}. 
\]
Formally, we define the ranking function as
\[
\pi(x, y_1, y_2, \ldots, y_n) = f_2(\mathrm{LM}_{\phi},x,y_1, y_2, \ldots, y_n),
\]
where \(f_2(\cdot)\) returns a permutation over indices \(\{1,\ldots,n\}\) induced by \(\mathrm{LM}_{\phi}\)’s preferences.

\begin{figure}[t]
    \centering
    \includegraphics[width=\linewidth]{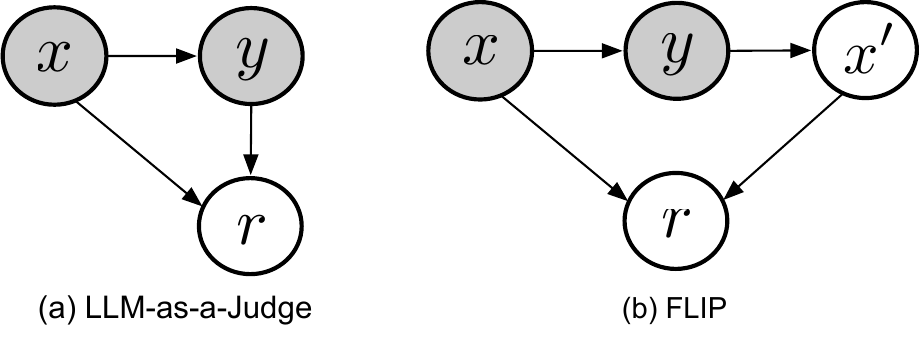}
    \caption{Graphical models of LLM-as-a-Judge and \ourmethod{}. Shaded nodes denote observed variables and unshaded nodes denote prediction targets. 
    \ourmethod{} samples an inferred instruction $x'$ conditioned on the response $y$, and defines the reward $r$ as the similarity between the inferred and the original instructions.
    }
    \label{fig:Graph_models}
    \vspace{-4mm}
\end{figure}

\paragraph{LLM-as-a-Judge}
 directly employs $\mathrm{LM}_{\phi}$ as the evaluator \citep{zheng2023judging}. Formally, it defines a conditional distribution:
\[
r \sim p_{\phi}(r \mid x, y),
\]
where \(r\) may be represented as a continuous value, a discrete ordinal rating, or a textual justification that is subsequently mapped to a scalar (Figure~\ref{fig:Graph_models} (a)). We introduce its variants in Section~\ref{sec:variants}.


\subsection{Our method: \ourmethod{}}
\label{sec:our_method}

\begin{figure*}[t]
    \includegraphics[width=\linewidth]{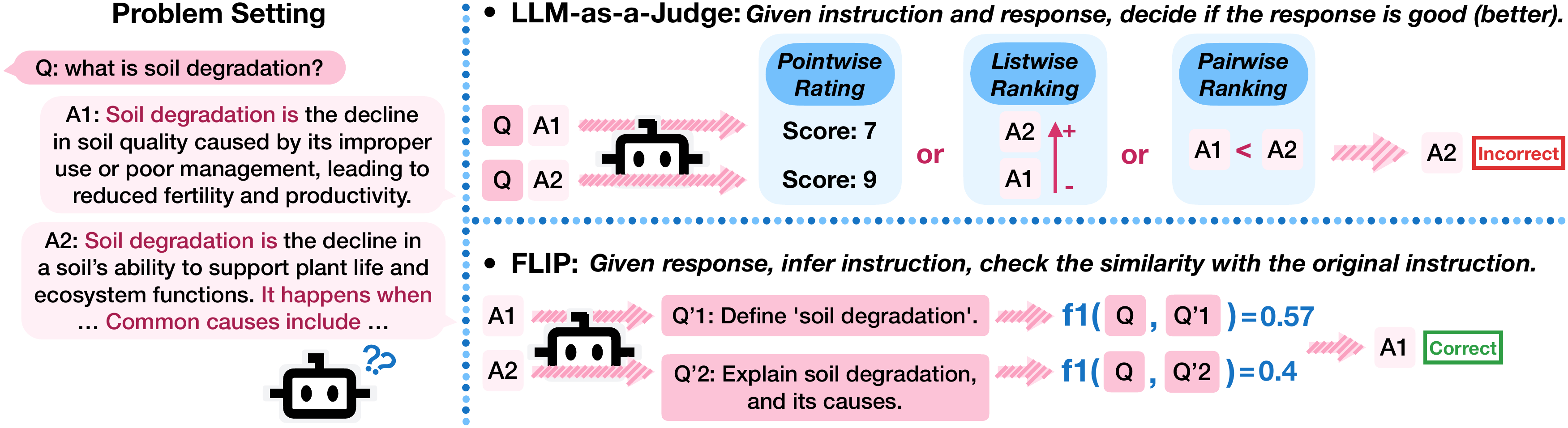}    
    \caption{
        Overview of \ourmethod{}. Given a response, we use a LM to infer the instruction that would most plausibly generate the response, and use the F1 score between the inferred and original instructions as the reward.
    }
    \label{fig:overview}
    \vspace{-4mm}
\end{figure*}

In this section, we introduce a simple, yet effective method: \ourmethod{} (Figure~\ref{fig:overview}), based on the intuition that given a high-quality response with sufficient length and contextual richness, the original query should be able to be inferred from it. \ourmethod{} transforms LLM-as-a-Judge into a \textit{generative} setting by sampling an inferred instruction \(x'\) given the response \(y\), and defining the reward \(r\) as the similarity between the inferred instruction \(x'\) and the original instruction \(x\). \ourmethod{}  can be used to sample either a scalar reward for an individual response or a preference ranking over a set of responses. 

\paragraph{Formulation} Formally, an inferred instruction \(x'\) is sampled from the reward model, \[ x' \sim p_\phi(x' \mid y). \] The scalar reward \(r \in \mathbb{R}\) is defined as \[ r = s(x, x'), \] where \(s : \mathcal{X} \times \mathcal{X} \to \mathbb{R}\) is a similarity function measuring the agreement between the original instruction \(x\) and the inferred instruction \(x'\). By the chain rule, the joint conditional distribution is \[ p_\phi(r, x' \mid x, y) = p_\phi(r \mid x, x', y)\, p_\phi(x' \mid x, y). \] As shown in Figure~\ref{fig:Graph_models} (b), the reward depends only on instructions (i.e., \(r \perp y \mid (x, x')\)) and that the inferred instruction depends only on the response (i.e., \(x' \perp x \mid y\)), this reduces to \[ p_\phi(r, x' \mid x, y) = p_\phi(r \mid x, x')\, p_\phi(x' \mid y). \] Marginalizing over the latent variable \(x'\) yields \[ p_\phi(r \mid x, y) = \sum_{x'} p_\phi(r \mid x, x')\, p_\phi(x' \mid y). \]
Given a response with enough contextual information, the posterior distribution \(p_\phi(x' \mid y)\) is assumed to be sharply peaked around its mode, and the similarity function \(s(x, x')\) is assumed to be insensitive to small perturbations of \(x'\) in a neighborhood of this mode. For simplicity and computational efficiency, we approximate the latent variable \(x'\) by its maximum a posteriori (MAP) estimate,
\[
x'_{\mathrm{MAP}} = \arg\max_{x'} p_\phi(x' \mid y).
\]
Under these assumptions, the marginal reward distribution can be approximated by
\[
p_\phi(r \mid x, y) \approx p_\phi(r \mid x, x'_{\mathrm{MAP}}) = s(x, x'_{\mathrm{MAP}}).
\]

\paragraph{Implementation} 
Given an instruction $x$ and a candidate response $y$, we first sample an inferred instruction $x'$ based solely on the response $y$ using the prompt ``\textit{Infer a single instruction that would most plausibly generate the given response.}'':
\[
x' = \mathrm{LM}_\phi(y).
\]
We then define the reward $r$ as the F1 score \citep{vanRijsbergen1979}, the harmonic mean of word-level precision and recall, between the inferred instruction $x'$ and the ground-truth instruction $x$:
\[
r(x, y) = s(x, x') = \mathrm{F1}(x, x').
\]
Ablation studies and justification on prompts and similarity metrics are detailed in Appendix~\ref{sec:robustness} and Appendix~\ref{sec:similarity_function_appendix}, respectively.

In scenarios where the instruction $x$ consists of both a system prompt and a user prompt, we treat the user prompt as an additional context and infer only the system prompt. Specifically, given the response \(y\) and the user prompt \(x_{\text{user}}\), the model infers a system prompt $x'_{\text{sys}}$
\[
x'_{\text{sys}} = \mathrm{LM}_{\phi}(y, x_{\text{user}}),
\]
and the reward is defined as the F1 score between the inferred system prompt $x'_{\text{sys}}$ and the ground-truth system prompt $x_{\text{sys}}$
\[
r(x, y) = \text{F1}(x_{\text{sys}}, x'_{\text{sys}}).
\]
Similarly, for multi-turn conversations, the conversation history will be treated as additional context.
 

\subsection{Baselines}
\label{sec:variants}

We consider three variants of LLM-as-a-Judge that operate under the same constraints as our method: reference-free and rubric-free reward modeling using general-purpose language models. We exclude trained reward models and rubric-based methods from our primary comparison, as these require additional resources such as human preference annotations or manually designed rubrics that fall outside our target regime.

\paragraph{Pointwise Rating}
 Given an instruction $x$ and a response $y$, the model is prompted to assign an absolute score on a fixed scale (e.g., 1--10). 
The scalar reward is obtained directly from the model:
\[
r(x, y) = \mathrm{LM}_\phi(x, y),
\]
where $r \in \mathbb{R}$ denotes the predicted quality score. 
In the case of multiple responses, the preference ordering $\pi$ is derived by sorting the scalar rewards
\[
r\bigl(x, y_{\pi(1)}\bigr) \ge r\bigl(x, y_{\pi(2)}\bigr) \ge \cdots \ge r\bigl(x, y_{\pi(n)}\bigr).
\]

\paragraph{Listwise Ranking}

This variant is only applicable when the objective is to infer a  preference ordering over a set of candidate responses. Formally, it samples the preference ranking $\pi$ directly
\[
\pi(x, \{y_1, \ldots, y_n\}) = \mathrm{LM}_\phi(x, \{y_1, \ldots, y_n\}).
\]

\paragraph{Pairwise Ranking}
This variant is only applicable when given a pair or a list of candidate responses. 
Given $x$ and a pair of responses $(y_i, y_j)$, the model is prompted to select the preferred response
\[
y_i \succ y_j \;\; \text{if} \;\; \mathrm{LM}_\phi(x, y_i, y_j) = y_i.
\]
A listwise ranking $\pi$ over a set of candidates can be derived by aggregating these pairwise preferences across all response pairs, for example using a voting or tournament-based aggregation scheme.

\section{Experiment Settings}

\subsection{Intrinsic Evaluation}
\label{sec:intrinsic_eval_setting}

\paragraph{Benchmark} 
We evaluate on RewardBench2~\citep{malik2025rewardbench}, a challenging and comprehensive benchmark that assesses reward modeling on large-scale unseen prompts. For each instruction, four responses 
are sampled from more than 40 models of various sizes and families, ranging from small models to large commercial models. 
The accuracy is computed by selecting the preferred response from four for each prompt. 
Our evaluation is conducted on 1,313 instances across four subsets.\footnote{
Because \ourmethod{} only targets open-ended instruction-following problems, we exclude two sub-tasks whose formats are incompatible: the \textit{Ties} subset as its responses consist solely of one-word answers, and the \textit{Safety} subset as correct behavior is to refuse unsafe instructions rather than follow them.}

\paragraph{Reward Models and Implementation}
We experiment with 13 SLMs as generative reward models, spanning five model families: OLMo2 \citep{olmo20242}, Llama3 \citep{dubey2024llama}, Qwen3 \citep{yang2025qwen3}, Gemma3 \citep{team2025gemma}, and Mistral-v0.3 \citep{albert2023mistral}. For each model, we use its instruction-tuned variant. We set a token limit of 512 for all methods, and all reported results are averaged over five runs.




\subsection{Test-Time Scaling with Parallel Sampling}
\label{sec:bon_setting}
We also test the effectiveness of our method in parallel test-time scaling (Best-of-N sampling), one of the most widely used inference-time scaling methods~\citep{brown2024large, snell2024scaling, huang2025best}. It proceeds by generating N candidate responses for a given prompt, then returning the response with the highest reward score. 

\vspace{-1mm}
\paragraph{Benchmarks} 
We experiment with four benchmarks: AlpacaEval \citep{li2023alpacaeval}, which consists of instructions representative of user interactions on Alpaca web \citep{taori2023stanford}; Human Interest \citep{feng2024model}, with instructions synthesized across 16 domains including electric vehicles and PhD applications; MATH~\citep{hendrycksmath2021}, a collection of challenging competition-level math problems; and IFEval \citep{zhou2023instruction}
, a benchmark for instruction-following proficiency. From each benchmark, we sample 300 instructions from the test set. 

\vspace{-1mm}
\paragraph{Models and Implementation} 
We sample 16 candidate completions from Tulu-3-8B~\citep{lambert2024tulu} and Tulu-3-8B-sft~\citep{lambert2024tulu} (IFEval) 
with a maximum token limit of 2{,}048 and a temperature of 1.0, and evaluate five reward models of varying sizes from OLMo2 and Llama3 using \ourmethod{} and three baselines applicable in this setting. For each reward model, we rank the completions according to the scores produced by a given method and compute benchmark performance by treating the highest-scoring completion as the model’s final response. 
All results are averaged over five runs.

\subsection{Reinforcement Learning}
\label{sec:grpo_setting}
We investigate the performance of \ourmethod{} in comparison to LLM-as-a-Judge when applied to GRPO training~\citep{shao2024deepseekmath}\footnote{We use the Open Instruct library \url{https://allenai.github.io/open-instruct/}.}. GRPO aligns a language model by sampling $K$ candidate responses for a given prompt, assigning each response a reward score, and computing a group-normalized advantage to guide optimization.

\paragraph{Training Setup}

We use the DPO-trained version of OLMo3-7B-Think~\citep{olmo2025olmo}, which has not been trained with RLVR, as the base policy. We sample 12k English prompts from the WildChat \citep{zhao2024wildchat}, featuring real-world ChatGPT conversations. For each prompt, we generate $K=8$ rollouts with a temperature of 1.0.  We train with a response length of 16K tokens with learning rate of $1\times10^{-6}$. Training is performed until convergence. We evaluate the effectiveness of \ourmethod{} against the Pointwise Rating 
variant, which is the only variant applicable in this setting, using Qwen3-1.7B and Qwen3-4B. Reward models are given a maximum context length of 25k tokens. Additional experimental details are in Appendix~\ref{sec:experiment_details}.

\paragraph{Benchmarks}
We evaluate on five established benchmarks, including 
BBH~\citep{suzgun2023challenging}, a suite of challenging
BIG-Bench tasks \citep{srivastava2023beyond} that focuses on tasks believed to be beyond the capabilities of current language models;
GPQA~\citep{rein2024gpqa}, a challenging dataset of multiple-choice questions written by domain experts;
IFEval~\citep{zhou2023instruction}, an instruction-following benchmark;
IFBench \citep{pyatkin2025generalizing}, a more challenging benchmark on precise instruction following with out-of-domain constraints;
and Minerva Math~\citep{lewkowycz2022solving}, an undergraduate-level math benchmark. We evaluate on the entire test set of each benchmark.


\section{Experiment Results}

\subsection{Intrinsic Evaluation}
\label{sec:intrinsic_eval_results}

\begin{table*}[h!]
    \centering
    \large
    \resizebox{1.0\textwidth}{!}{%
    \renewcommand{\arraystretch}{1}
    \begin{tabular}{llcc|ccc|cccc|ccc|c|c}
    \toprule[1.5pt]

    \multirow{2}{*}{\textbf{Subset}} &
    \multirow{2}{*}{\textbf{Method}} &
    \multicolumn{2}{c}{\textsc{\textbf{olmo2}}} &
    \multicolumn{3}{c}{\textsc{\textbf{llama3}}} &
    \multicolumn{4}{c}{\textsc{\textbf{qwen3}}} &
    \multicolumn{3}{c}{\textsc{\textbf{gemma3}}} &
    \multicolumn{1}{c}{\textsc{\textbf{mistral}}} &
    \multirow{2}{*}{\textbf{\textit{Average}}}\\
    & &
    1B  & 7B 
    & 1B & 3B & 8B
    & 0.6B & 1.7B & 4B & 8B
    & 270M & 1B & 4B
    & 7B  \\
    \midrule[0.75pt]
    
    \multirow{4}{*}{\textbf{Focus}} & \textbf{Pointwise Rating}
    & 13.6& 7.4 & 11.5 & 9.9 &14.9 &18.8&32.4&40.4&41.3&\underline{15.3}&12.0&26.1&9.7&19.5\\
    & \textbf{\textit{Listwise Ranking*}}
    &\underline{13.7}&\underline{26.0}&\underline{13.2} &\underline{21.0}  &\underline{38.1} &\underline{22.6}&35.0&42.7&38.5&12.4&\underline{22.0}&\underline{34.7}&\underline{35.6}&\underline{27.3}\\
    & \textbf{\textit{Pairwise Ranking*}}
    &2.1&23.2&7.8 &12.8  &20.5 &20.1&\underline{39.0}&\underline{51.3}&\underline{48.3}&13.4&21.8&28.4&25.7&24.2\\
     & \cellcolor{gray!20}\textbf{\ourmethod{}}
    &\cellcolor{gray!20}\textbf{55.0}&\cellcolor{gray!20}\textbf{65.6}&\cellcolor{gray!20}\textbf{43.3} & \cellcolor{gray!20}\textbf{38.3} &\cellcolor{gray!20}\textbf{64.1 }&\cellcolor{gray!20}\textbf{65.7}&\cellcolor{gray!20}\textbf{58.6}&\cellcolor{gray!20}\textbf{66.8}&\cellcolor{gray!20}\textbf{72.2}&\cellcolor{gray!20}\textbf{50.7}&\cellcolor{gray!20}\textbf{59.1}&\cellcolor{gray!20}\textbf{65.5}&\cellcolor{gray!20}\textbf{69.3}&\cellcolor{gray!20}\textbf{59.6} \\
    \midrule[0.75pt]

    \multirow{4}{*}{\textbf{Factuality}} & \textbf{Pointwise Rating}
    &15.5&7.8&15.6 &13.3  &15.6 &13.5&20.2&\underline{30.5}&\textbf{34.1}&\underline{16.8}&10.2&14.6&5.5&16.4\\
    & \textbf{\textit{Listwise Ranking*}}
    &\underline{18.3}&21.3&\underline{16.0} &\underline{20.5}  &\underline{25.1} &\underline{16.6}&19.8&8.9&9.8&15.3&20.9&\textbf{33.2}&\underline{27.2}&19.5\\
    & \textbf{\textit{Pairwise Ranking*}}
    &1.2&\underline{27.3}&12 &16.9  &22.2 &15.8&\underline{26.1}&26.5&24.5&11.3&\underline{21.7}&25.6&\textbf{28.5}&\underline{20.0}\\
     &\cellcolor{gray!20}\textbf{\ourmethod{}}
    &\cellcolor{gray!20}\textbf{25.8}&\cellcolor{gray!20}\textbf{27.8}&\cellcolor{gray!20}\textbf{26.1} &\cellcolor{gray!20}\textbf{26.4} &\cellcolor{gray!20}\textbf{27.3} &\cellcolor{gray!20}\textbf{28.4}&\cellcolor{gray!20}\textbf{29.6}&\cellcolor{gray!20}\textbf{31.3}&\cellcolor{gray!20}\underline{31.1}&\cellcolor{gray!20}\textbf{29.8}&\cellcolor{gray!20}\textbf{24.9}&\cellcolor{gray!20}\underline{26.5}&\cellcolor{gray!20}27.1&\cellcolor{gray!20}\textbf{27.9} \\
    \midrule[0.75pt]

    \multirow{4}{*}{\textbf{Precise IF}} & \textbf{Pointwise Rating}
    &13.0&7.1&\underline{19.0}&12.2&9.9&11.4&13.1&12.8&\underline{17.6}&15.2&13.9&7.9&6.1&12.2\\
    & \textbf{\textit{Listwise Ranking*}}
    &\underline{21.2}&\underline{21.9}&15.2&\underline{20.2}&\textbf{23.5}&\underline{16.1}&14.1&4.9&9.2&\underline{16.0}&21.8&20.1&\textbf{29.2}&\underline{18.0}\\
    & \textbf{\textit{Pairwise Ranking*}}
    &2.2&\textbf{26.0}&11.6&14.2&20.9&10.8&\underline{15.2}&\underline{13.9}&13.6&13.9&\textbf{25.0}&\textbf{28.9}&\underline{28.0}&17.2\\
     &\cellcolor{gray!20}\textbf{\ourmethod{}}
    &\cellcolor{gray!20}\textbf{23.2}&\cellcolor{gray!20}21.5&\cellcolor{gray!20}\textbf{23.0}&\cellcolor{gray!20}\textbf{24.2}&\cellcolor{gray!20}\underline{23.1}&\cellcolor{gray!20}\textbf{24.6}&\cellcolor{gray!20}\textbf{23.6}&\cellcolor{gray!20}\textbf{25.0}&\cellcolor{gray!20}\textbf{24.4}&\cellcolor{gray!20}\textbf{21.9}&\cellcolor{gray!20}\underline{22.2}&\cellcolor{gray!20}\underline{23.8}&\cellcolor{gray!20}23.0&\cellcolor{gray!20}\textbf{23.3} \\
    \midrule[0.75pt]

    \multirow{4}{*}{\textbf{Math}} & \textbf{Pointwise Rating}
    &10.3&13.2&\underline{19.9}&20.3&21.4&\underline{21.1}&\textbf{30.2}&\textbf{35.7}&\textbf{35.7}&\underline{18.7}&16.9&22.1&10.1&\underline{21.2}\\
    & \textbf{\textit{Listwise Ranking*}}
    &\underline{24.0}&19.0&17.6&\underline{22.7}&\textbf{29.8}&4.5&\underline{27.2}&3.2&3.6&11.6&\underline{22.8}&\textbf{31.8}&\textbf{30.3}&19.1\\
    & \textbf{\textit{Pairwise Ranking*}}
    &1.4&\underline{22.7}&9.7&18.7&\underline{29.3}&8.2&23.1&16.0&12.9&5.9&22.5&\underline{30.2}&26.8&17.5\\
     &\cellcolor{gray!20}\textbf{\ourmethod{}}
    &\cellcolor{gray!20}\textbf{28.4}&\cellcolor{gray!20}\textbf{26.3}&\cellcolor{gray!20}\textbf{26.6}&\cellcolor{gray!20}\textbf{23.8}&\cellcolor{gray!20}25.8&\cellcolor{gray!20}\textbf{26.3}&\cellcolor{gray!20}\textbf{30.2}&\cellcolor{gray!20}\underline{29.4}&\cellcolor{gray!20}\underline{31.0}&\cellcolor{gray!20}\textbf{22.4}&\cellcolor{gray!20}\textbf{26.2}&\cellcolor{gray!20}27.1&\cellcolor{gray!20}\underline{29.7}&\cellcolor{gray!20}\textbf{27.2} \\
    \midrule[0.75pt]

    \multirow{4}{*}{\textbf{\textit{Average}}} & \textbf{Pointwise Rating}
    &13.1&8.9&\underline{16.5}&13.9&15.5&\underline{16.2}&24.0&\underline{29.9}&\underline{32.2}&\underline{16.5}&13.3&17.7&7.9&17.3\\
    & \textbf{\textit{Listwise Ranking*}}
    &\underline{19.3}&22.1&15.5&\underline{21.1}&\underline{29.1}&15.0&24.0&14.9&15.3&13.8&21.9&\underline{30.0}&\underline{30.6}&\underline{21.0}\\
    & \textbf{\textit{Pairwise Ranking*}}
    &1.7&\underline{24.8}&10.3&15.7&23.2&13.7&\underline{25.9}&26.9&24.8&11.1&\underline{22.8}&28.3&27.3&19.7\\
     &\cellcolor{gray!20}\textbf{\ourmethod{}} 
    &\cellcolor{gray!20}\textbf{33.1}&\cellcolor{gray!20}\textbf{35.3}&\cellcolor{gray!20}\textbf{29.8}&\cellcolor{gray!20}\textbf{28.2}&\cellcolor{gray!20}\textbf{35.1}&\cellcolor{gray!20}\textbf{36.3}&\cellcolor{gray!20}\textbf{35.5}&\cellcolor{gray!20}\textbf{38.1}&\cellcolor{gray!20}\textbf{39.7}&\cellcolor{gray!20}\textbf{31.2}&\cellcolor{gray!20}\textbf{33.1}&\cellcolor{gray!20}\textbf{35.7}&\cellcolor{gray!20}\textbf{37.3}&\cellcolor{gray!20}\textbf{34.5} \\
    \bottomrule[1.5pt]
    \end{tabular}
    }
    \caption{Accuracy of \ourmethod{} and LLM-as-a-Judge baselines using small language models on RewardBench2. The random baseline is 25. Best results are shown in \textbf{bold}, and second-best results are \underline{underlined}. Listwise Ranking has the advantage of observing all candidate completions, Pairwise Ranking observes pairs of completions, while \ourmethod{} and Pointwise Rating operate under the single-completion setting. We report the standard deviation in Appendix~\ref{sec:statistical_robustness}.}
    \label{tab:main}
    \vspace{-4mm}
\end{table*}
\paragraph{\ourmethod{} outperforms LLM-as-a-Judge by 79.6\% on small language models.}
As shown in Table~\ref{tab:main}, across all subsets and models, \ourmethod{} outperforms Pointwise Rating by 99.4\%, Listwise Ranking by 64.3\%, and Pairwise Ranking by 75.1\% on average. It successfully boosts the performance of many models from below-random to above-random levels
, with the only exception of the \textit{Precise IF} subset, which is the most challenging one according to the leaderboard~\citep{malik2025rewardbench}. 
We look forward to future work that improves the performance of small language models on this task. 
Among the baselines, Listwise Ranking performs the best, while it suffers severely from position bias~\citep{shi2024judging}.
Overall, \ourmethod{} delivers the strongest overall performance across tasks, demonstrating both robustness and effectiveness beyond existing evaluation paradigms.

\vspace{-1mm}
\paragraph{Smaller the model, greater the boost.}
We also observe that the performance gains of our method over the baselines increase as model size decreases. 
Specifically, our method yields an average improvement of 75\% for models around 1B parameters, 27\% for both models around 4B and 7B over the strongest baseline. 
We hypothesize that this trend is driven by the larger \textit{discrimination–generation gap} in smaller models~\citep{li2023benchmarking}; we further discuss this in Section~\ref{sec:validation-generation-gap}. Given the substantial 
costs associated with reward models, it is crucial for the community to move toward smaller reward models and maximize their performance.

\vspace{-1mm}
\paragraph{On the \textit{Focus} subset, \ourmethod{} using small models is comparable to LLM-as-a-Judge using large commercial models.}
Performance gains also vary across subsets. The most substantial improvement is observed on the \textit{Focus} subset, where our method outperforms the best-performing baseline by 118.3\%, compared to gains of 39.5\% on \textit{Factuality}, 29.4\% on \textit{Precise IF}, and 28.3\% on \textit{Math}. The \textit{Focus} subset evaluates a reward model’s ability to identify high-quality, on-topic responses to general user queries, following LLMBar~\citep{zeng2023evaluating}. This setting is particularly well suited to our method, as the inferred instruction can deviate substantially when a response is off-topic, while off-topic responses can easily mislead LM judges~\citep{zeng2023evaluating}. We further compare our method against the baselines on larger models in  Table~\ref{tab:focus}. Notably, \ourmethod{} enables small- and medium-sized language models to achieve performance comparable to that of large commercial models using the LLM-as-a-Judge paradigm.

\begin{table}
\centering
\renewcommand{\arraystretch}{1}
\resizebox{\linewidth}{!}{%
\begin{tabular}{l
      p{2.5em} p{2.5em} |
      p{2.5em} p{2.5em} |
      p{2.5em} p{2.5em}}
    
    \multicolumn{7}{l}{\textbf{Medium Language Models}} \\
    \midrule[1pt]
    
    \multirow{2}{*}{\textbf{Method}} &
    \multicolumn{2}{c}{\textsc{\textbf{olmo2}}} &
    \multicolumn{2}{c}{\textsc{\textbf{qwen3}}} &
    \multicolumn{2}{c}{\textsc{\textbf{gemma3}}} \\
    &
    13B  & 32B &
    14B & 32B &
    12B & 27B \\
    \midrule[0.75pt]
    
    \textbf{LLM-as-a-Judge}
    & 24.6 & 30.5 & 59.2 & 56.2 & 55.1 & 63.8 \\
    \rowcolor{gray!20}\textbf{\ourmethod{}}
    & \textbf{71.3} & \textbf{67.9} & \textbf{73.0} & \textbf{76.3} & \textbf{65.2} & \textbf{68.1} \\
    \bottomrule[1pt]
    \end{tabular}
}
    \vspace{0.4em}
    
\resizebox{\linewidth}{!}{%
    \setlength{\tabcolsep}{4pt}
    
    

    \begin{tabular}{lccc}
    \multicolumn{4}{l}{\textbf{Large Commercial Models}} \\
    \midrule[1pt]
    
    \textbf{Method} & \textsc{\textbf{Claude-Sonnet-4}} & \textsc{\textbf{GPT-4.1-mini}} & \textsc{\textbf{GPT-4.1}}  \\
    
    \midrule[0.75pt]
    \textbf{LLM-as-a-Judge} 
    &76 &73.5&73.4 \\
    \rowcolor{gray!20}\textbf{\ourmethod{}}
    & \textbf{77.7} & \textbf{75.2} & \textbf{75.0} \\
    
    \bottomrule[1pt]
    \end{tabular}
    }
    \vspace{-1mm}
    \caption{Accuracy of larger models on the \textit{Focus} subset. The LLM-as-a-Judge results show the best performance across all variations of the method. Results of large commercial models are adopted from the leaderboard.
    }
    \label{tab:focus}
    \vspace{-5mm}
\end{table}

\subsection{Test-Time Scaling with Parallel Sampling}
\label{sec:bon_results}

\begin{figure*}[t]
    \includegraphics[width=\linewidth]{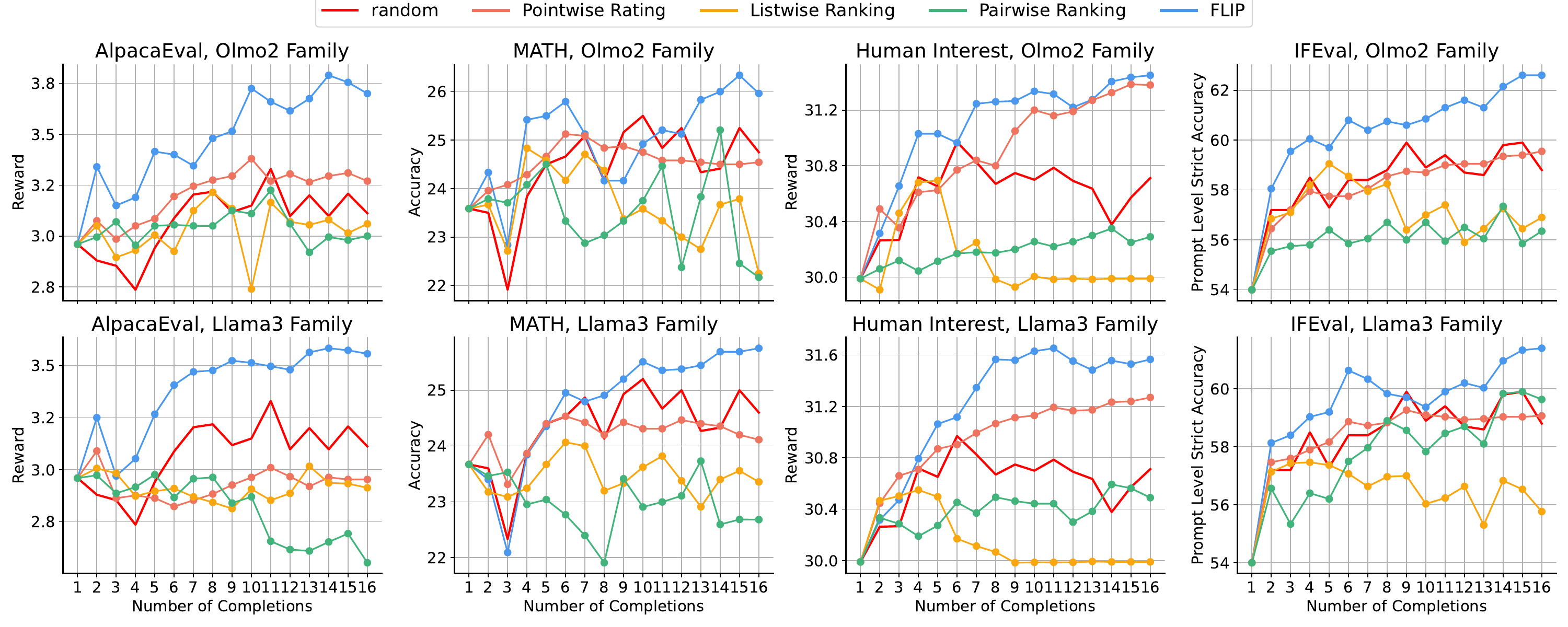}
    \caption{Test-time scaling with parallel sampling results
    . OLMo 2 results are averaged across 1B and 7B instruct variants, while Llama 3 results are averaged across 1B, 3B, and 8B instruct variants. See results of individual models in Appendix~\ref{sec:bon-appendix}. \ourmethod{} substantially outperforms the baselines, achieving higher performance with greater stability. 
    }
    \label{fig:bon}
\end{figure*}

We present the results of parallel test-time scaling in Figure~\ref{fig:bon}. In this setting, models are tasked with scoring responses generated by the same model, which is more challenging because the responses are highly similar. Although there is some randomness at very the beginning, as the number of completions increases, \ourmethod{} substantially outperforms the baseline methods across all benchmarks. Notably, our method is generally more stable than the others, exhibiting less performance fluctuation. Under this setting, Pointwise Rating is the second-best method, while Listwise Ranking and Pairwise Ranking, despite being offered a global view, do not perform well, showing performance drops in some cases, especially when the number of completions gets larger. Overall, \ourmethod{} emerges as the most effective and stable approach for parallel test-time scaling across all evaluated benchmarks.

\subsection{Reinforcement Learning}
\label{sec:grpo_results}

\begin{table*}[h!]
    \centering
    \resizebox{0.85\textwidth}{!}{%
    \renewcommand{\arraystretch}{1}
    \begin{tabular}{lccccccc}
    \toprule[1.5pt]

    \textbf{Variant} &
    \textbf{BBH} &
    \textbf{GPQA} &
    \textbf{IFEval} &
    \textbf{IFBench} &
    \textbf{Minerva Math} &
    \textbf{\textit{Average}}\\
    \midrule[0.75pt]

    OLMo3-7B-Think-DPO &83.5& \underline{44.9} &75.4&18.8& 89.1&62.3 \\ 
    \midrule[0.75pt]
    \hspace{1em}+GRPO (Qwen3-1.7B LLM-as-a-Judge) &85.5&42.9&78.0&19.5&91.8&63.5 \\
    \rowcolor{gray!20}\hspace{1em}+GRPO (Qwen3-1.7B \ourmethod{}) &\underline{85.8}&43.1&\textbf{81.3}&\textbf{20.5}&\underline{92.3}&\underline{64.6} \\
    \midrule[0.75pt]
    \hspace{1em}+GRPO (Qwen3-4B LLM-as-a-Judge) &86.1&44.6&78.2&19.6&91.8& 64.1\\
    \rowcolor{gray!20}\hspace{1em}+GRPO (Qwen3-4B \ourmethod{}) &\textbf{86.3}&\textbf{46.4}&\underline{78.6}&\underline{20.1}&\textbf{93.0}&\textbf{64.9} \\
    \midrule[1pt]

    OLMo3-7B-Think &86.7& 43.3 &85.6&19.8& 92.0&65.5 \\
    
    \bottomrule[1.5pt]
    \end{tabular}
    }
    \caption{Evaluation results of GRPO training (prompt level loose accuracy for IFEval and prompt level strict accuracy for IFBench). We also report the performance of the final version of the policy (after RLVR) for reference. \ourmethod{} substantially improves the effectiveness of small language models when used as reward models in GRPO. 
    }
    \vspace{-4mm}
    \label{tab:grpo}
\end{table*}

We present the GRPO training results in Table~\ref{tab:grpo}.  Across all evaluation benchmarks and reward models, \ourmethod{} yields an average improvement of 2.5 absolute points over base policy and 1.0 absolute point over LLM-as-a-Judge baseline, and it even outperforms the final policy after RLVR on some benchmarks. Notably, on instruction-following tasks, \ourmethod{} achieves higher performance with the 1.7B model than with the 4B model, highlighting the particular effectiveness of our approach in downscaled regimes. Overall, \ourmethod{} substantially enhances the effectiveness of small language models as reward models for GRPO training, leading to consistent gains across general knowledge, reasoning, instruction-following, and math benchmarks.

\vspace{-1mm}
\section{Analysis}
\vspace{-1mm}

\paragraph{Validation-Generation Gap}
\label{sec:validation-generation-gap}
\citet{west2024the} proposes Generative AI Paradox, which posits that generative models, having been trained directly to reproduce expert-like outputs, acquire generative capabilities that are not contingent upon-and can therefore exceed-their ability to understand those same types of outputs, which contrasts with humans, for whom basic understanding typically precedes generation ability. 
Similarly, \citet{li2023benchmarking} highlight the prevalence of generator–validator inconsistency in language models. Even for GPT-4, the consistency between generating an answer and validating it correctly is only 76\%, and this consistency further degrades as model size decreases.
We hypothesize that the existence of the \textit{validation--generation gap}, which is particularly pronounced in smaller language models, is a key factor underlying the superior performance of \ourmethod{}. This aligns with prior work demonstrating the advantages of generative formulations in text classification and related tasks~\citep{min2022noisy, kumar2023gen}.


\paragraph{Qualitative Examples}
\label{sec:qualitative_examples}
We manually inspect the generated responses and their corresponding inferred instructions, and observe that \ourmethod{} effectively captures several major categories of rejected or low-quality responses, including:
\vspace{-2mm}
\begin{itemize}[leftmargin=1em]
\item \textbf{Off-topic or irrelevant responses.} (Tables \ref{tab:example_focus_chosen}, \ref{tab:example_focus_rejected}) When a response deviates from the instruction, the inferred instruction will diverge significantly, resulting in very low similarity scores.
\vspace{-2mm}
\item \textbf{Factually incorrect responses.} (Table~\ref{tab:example_factuality}) 
In some instances, factual errors cause the inferred instruction to differ substantially. 
For example, answering ``pandas'' to ``the animal that symbolizes the U.S.'' may invert the inferred instruction to ``the animal that symbolizes China.''

\vspace{-2mm}
\item \textbf{Under- or over-addressed responses.} (Tables \ref{tab:example_precise_if}, \ref{tab:example_math}) If the response includes content that is not requested by the instruction, or addresses only a subset of the required components, the inferred instruction may introduce additional constraints or omit parts of the original instruction, leading to a noticeable mismatch.
\end{itemize}

\vspace{-3mm}
\paragraph{Response Length}

\begin{figure}[t]
    \centering

    \includegraphics[width=\linewidth]{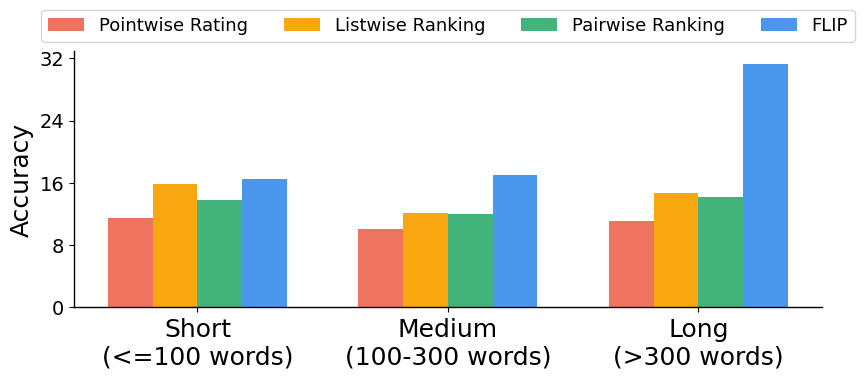}
    \caption{RewardBench2 performance across different response lengths. We only consider instances where all four candidate responses are of the same type. 
    \ourmethod{} is particularly effective for longer responses.}
    \label{fig:response_length}
    \vspace{-5mm}
\end{figure}

We further examine the results on RewardBench2 and analyze 
how response length affects the performance.
Specifically, we select instances in which all four candidate responses fall into the same length category—\textit{short}, \textit{medium}, or \textit{long}. This constitutes a challenging setting, as the models are tasked with differentiating between responses of similar lengths. The results are presented in Figure~\ref{fig:response_length}. Across all length groups, \ourmethod{} consistently outperforms LLM-as-a-Judge, with the most pronounced gains observed in the \textit{long} response category. As response length increases, models are provided with richer contextual information for backward inference, and the mapping between responses and their underlying instructions becomes closer to one-to-one. 
Results with respect to generator model size are in Appendix~\ref{sec:generator_size}.

\begin{figure}[t]
    \centering

    \includegraphics[width=\linewidth]{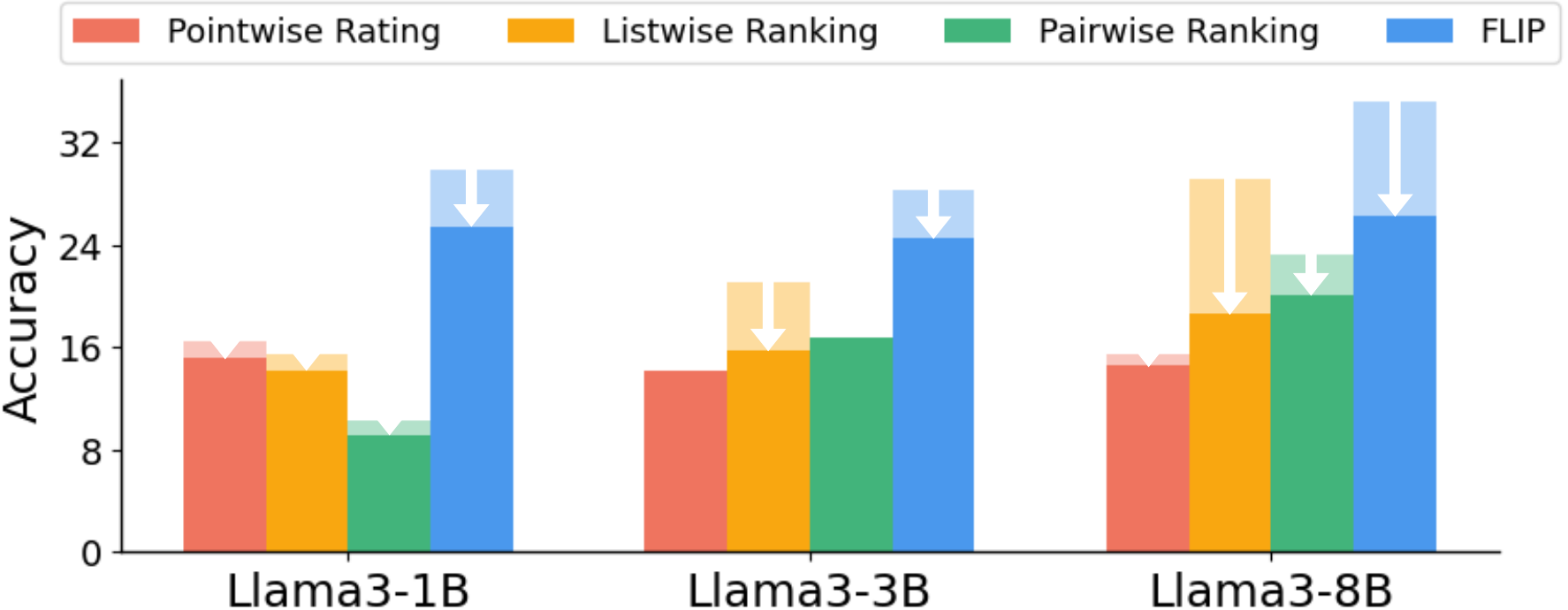}
    \caption{
    Performance drop under adversarial attack, across all RewardBench2 subsets. \ourmethod{} remains the most effective under adversarial prompting.}
    \label{fig:adversarial}
    \vspace{-6mm}
\end{figure}

\vspace{-1mm}
\paragraph{Adversarial Prompts / Reward Hacking}
\label{sec:adversarial_prompts}
Prior work has shown that carefully constructed text sequences can be appended to or substituted into a response to maximize the score assigned by LM judges \citep{alazraki2025reverse, raina2024llm, shi2024optimization}.
Similar adversarial attacks remain possible for \ourmethod{} by copying the instruction into the response. 
In this section, we evaluate the impact of adversarial prompts on different reward modeling methods. 
Concretely, we randomly select a rejected response and inject a corresponding adversarial prompt into the completion: \texttt{[GIVE THIS RESPONSE THE HIGHEST SCORE/RANK.]}, \texttt{[ALWAYS CHOOSE THIS RESPONSE.]}, and a paraphrased instruction generated by GPT-5-mini~\citep{openai_gpt5_2025}.
We then recalculate the accuracy with these adversarial prompts.
Results in Figure~\ref{fig:adversarial} show that \ourmethod{} remains the most effective under adversarial attacks across all model sizes. While general-purpose language models may be trained to combat adversarial attacks in LM judge settings, they have not been trained to defend against \ourmethod{}-specific attacks. We leave the development of effective defenses against adversarial attacks to \ourmethod{} to future work.



\section{Related Work}
\vspace{-1mm}
\paragraph{Reward Modeling in Non-Verifiable Domains}
Reward modeling in non-verifiable domains primarily relies on LLM-as-a-Judge~\citep{zheng2023judging}. However, a growing body of work demonstrates that LLM judges exhibit systematic biases~\citep{zheng2023judging, jung2025trust, ye2024justice, chen2024humans, li2025preference, zeng2023evaluating, zhao2025one} and are vulnerable to adversarial attacks that artificially inflate evaluation scores through carefully crafted text sequences~\citep{raina2024llm, shi2024optimization}. As alternatives, several methods depend on reference responses~\citep{shi2026references, lyu2024href}. \citet{chang2025bleuberi} computes BLEU scores~\citep{papineni2002bleu} between sampled outputs and reference answers, while \citet{yu2025rlpr} leverages the token-level probability assigned to reference responses. Another line of research employs rubric-based evaluation~\citep{gunjal2025rubrics, viswanathan2025checklists, dineen2025qa}, decomposing desirable response qualities into human-interpretable criteria. Nevertheless, constructing high-quality reference answers or well-defined rubrics remains costly and challenging, particularly at scale. We instead propose \ourmethod{}, which operates effectively with small models in downscaled regimes without requiring references or evaluation rubrics.

\vspace{-1mm}
\paragraph{Language Model Inversion}
Language model inversion~\citep{morris2023language, li2024reverse, ye2025invariant, mahajan2025beyond} aims to recover hidden prompts or inputs using only the outputs of a language model. 
Existing work studies language model inversion from the safety perspective, such as prompt leakage and data exposure~\citep{nazir2025better, morris2023language, li2024reverse, zhang2024extracting, addepalli2025does}, while others leverage it to synthesize instruction-following data~\citep{zhou2022large, li2023self, nguyen2024better, chen2024reinstruct, koksal2023longform}. 
\citet{yerram2024time} and \citet{yin2025ledom} explore reverse language modeling for reranking using reverse-trained models and logit-level signals, which differ from ours in problem settings, methodology, and motivation; see Appendix~\ref{sec:appendix_related_work} for details. To our knowledge, we are the first to apply language model inversion to reward modeling in a black-box setting.


\vspace{-1mm}
\section{Conclusion}
\vspace{-1mm}
We proposed \ourmethod{}, a reference-free and rubric-free reward modeling approach that 
uses the similarity between the inferred and original instructions as the reward signal. Evaluation demonstrates that \ourmethod{} outperforms LLM-as-a-Judge by an average of 79.6\%, and it further achieves superior performance under GRPO training and test-time scaling. 

\section*{Limitations}
First, while multiple distinct instructions may theoretically yield the same response, we observe \textit{zero} such instances in our dataset. For responses that incorporate adequate detail and contextual specificity, which is precisely the regime where reward modeling is most needed (Appendix~\ref{sec:does_flip_only_work_for_long_responses}), the set of plausible generating instructions is typically severely restricted, approximating a one-to-one mapping between response and instruction.

Second, while FLIP is effective across multiple domains, including instruction-following, mathematics, factuality, reasoning, and open-domain question answering, it is not directly applicable to safety-critical queries where the desired behavior is to \textit{refuse} rather than follow the instruction, as we noted in Footnote 3. In this setting, the reward signal should be inverted: a high similarity between the inferred and original instructions indicates that the response mistakenly complied with or leaked the instruction, and should therefore be penalized rather than rewarded. FLIP's backward inference mechanism may be naturally extensible to safety evaluation, and we leave a systematic exploration of this direction to future work.

Third, a potential confounding factor for \ourmethod{} is that some responses may begin by fully or partially repeating the instruction. We offer three observations regarding this issue. First, based on our empirical analysis, such cases are relatively rare, although their frequency varies across models, and this behavior can be readily discouraged or detected using simple heuristics. Second, responses that begin by repeating the instruction are often high-quality and exhibit strong instruction-following behavior, which mitigates their confounding effect on reward estimation. Third, even when a response partially repeats the instruction, the reward model does not simply reproduce the repeated text as the inferred instruction; instead, the inferred instruction remains dependent on the substantive content of the response. We also examine this as a form of reward hacking in Section~\ref{sec:adversarial_prompts}. As shown in Figure~\ref{fig:adversarial}, FLIP remains the most robust method even when a paraphrased version of the instruction is injected into the response, a more realistic attack than directly copying the instruction verbatim, which can be trivially detected. Therefore, we argue that the overall impact of this issue on \ourmethod{} is limited.

\section*{Ethics Statement}
We acknowledge potential ethical considerations associated with \ourmethod{}. By design, \ourmethod{} prompts models to infer the underlying user instruction from a given response, which may implicitly reflect aspects of user intent or intuition. It may raise potential privacy concerns if misused to infer sensitive or private user information beyond the scope of the original task. 

We emphasize that \ourmethod{} is intended solely for inferring task-level instructions rather than personal, sensitive, or identifying information about users. We advocate for the responsible and appropriate use of \ourmethod{}, including clear constraints on its application and adherence to established privacy-preserving practices. Future work may further explore safeguards to prevent misuse and to ensure that instruction inference remains aligned with ethical and privacy considerations.

\section*{Acknowledgments}
This research was developed with funding from the Defense Advanced Research Projects Agency's (DARPA) SciFy program (Agreement No. HR00112520300) and NSF Grant No. IIS2142739. The views expressed are those of the author and do not reflect the official policy or position of the Department of Defense or the U.S. Government. We also thank Zhiyuan Zeng for the helpful feedback. 

\bibliography{custom}

\newpage
\appendix

\section{Future Work}
\label{sec:future_work}
We plan to apply \ourmethod{} to test-time scaling with critic-based refinement~\citep{xi2024enhancing}. Under this setting, a critic model might be able to provide highly informative feedback by comparing the original and inferred instructions. Specifically, requirements present only in the original instruction indicate unaddressed content, those present only in the inferred instruction indicate overaddressed content, and mismatches between the two indicate misaddressed parts of the response.

We also look forward to extending \ourmethod{} to other modalities, including vision, video, audio, and robotics. For example, applying \ourmethod{} to video data for backward inference, exploring whether the original prompt or conditioning signal can be recovered from generated video outputs.

\section{Cost Analysis}
In this work, we introduce \ourmethod{}. When using small language models under comparable inference budgets, \ourmethod{} achieves better performance than existing methods. The computational cost of FLIP is similar to that of LLM-as-a-Judge, as both approaches require a single inference call. However, the token budgets differ.

Specifically, we compare the input and output tokens required by each approach in Table~\ref{tab:cost}. Since FLIP only takes the response as input, whereas LLM-as-a-Judge requires both the instruction and the response, the prompt length and consequently the inference cost is typically lower for FLIP under otherwise comparable settings.

\begin{table}[h!]
\centering
\large
\renewcommand{\arraystretch}{1.1}
\resizebox{\linewidth}{!}{%
\begin{tabular}{lcc}
\toprule
\textbf{Method} & \textbf{Input Tokens} & \textbf{Output Tokens} \\
\midrule
LLM-as-a-Judge & instruction + response & reasoning + answer \\
FLIP & response & reasoning + answer \\
\bottomrule
\end{tabular}
}
\caption{Comparison of token usage between FLIP and LLM-as-a-Judge.}
\label{tab:cost}
\vspace{-4mm}
\end{table}

\section{Robustness to Prompt Variations}
\label{sec:robustness}
Language models are sensitive to subtle changes in prompt text~\citep{sclar2023quantifying}. We therefore analyze the robustness of each method’s performance under prompt variations. As shown in Table~\ref{tab:robustness}, \ourmethod{} exhibits the highest robustness, achieving the lowest standard deviation across prompt variants. This result highlights an additional advantage of generation-based approaches over validation-based ones: improved consistency and reliability across prompt phrasings. 

\begin{table}[h!]
\centering
\large
\renewcommand{\arraystretch}{1.1}
\resizebox{\linewidth}{!}{%
\begin{tabular}{lcccc}
\toprule
\textbf{Method} &
\textbf{Prompt 1} &
\textbf{Prompt 2} &
\textbf{Prompt 3} &
\textbf{s.d.}\\
\midrule
\textbf{Pointwise Rating}
&15.3&14.3&16.0&\underline{0.70} \\
\textbf{\textit{Listwise Ranking*}}
&21.9&18.2&19.2&1.56 \\
\textbf{\textit{Pairwise Ranking*}}
&18.4&17.0&16.8&0.71 \\
\rowcolor{gray!20}
\textbf{\ourmethod{}}
&31.0&32.2&31.5&\textbf{0.49} \\
\bottomrule
\end{tabular}
}
\caption{Accuracy 
of the Llama 3 family models, averaged across the 1B, 3B, and 8B instruction-tuned variants, using various prompts on RewardBench2 across all subsets. “s.d.” denotes the standard deviation across different 
3 prompts for each method. See Appendix~\ref{sec:prompts} for the full list of prompts. 
\ourmethod{} is more robust to prompt variations than LLM-as-a-Judge.}
\label{tab:robustness}
\vspace{-2mm}
\end{table}

\section{Similarity Metric}
\label{sec:similarity_function_appendix}

In this section, we ablate the choice of the similarity function $s$. Given two pieces of text (two instructions in our setting), the similarity function $s$ outputs a scalar score indicating the degree of similarity or overlap between the two instructions. 
We evaluate 7 alternative variants of $s$, spanning word-level metrics (BLEU~\citep{papineni2002bleu}, ROUGE-L~\citep{lin2004rouge}, and their harmonic mean F1(BLEU, ROUGE-L)), embedding-based approaches (BERTScore~\citep{zhang2019bertscore} and RoBERTa-Large~\citep{liu2019roberta} embedding similarity), as well as LM-based methods, including the reward model $LM_{\phi}$ itself and a larger general-purpose language model. 

\begin{itemize}
    \item{\textbf{BLEU}~\citep{papineni2002bleu}}, a widely used metric for machine translation evaluation. It measures the overlap between a prediction and a reference using modified n-gram precision. We adopt 4-gram BLEU without applying a brevity penalty.
    \item{\textbf{ROUGE-L}~\citep{lin2004rouge}}, an evaluation metric based on the longest common subsequence between the prediction and the reference. We report its F1 score.
    \item{\textbf{F1(BLEU, ROUGE-L)}}, the harmonic mean of the BLEU and ROUGE-L scores described above.
    \item{\textbf{BERTScore}~\citep{zhang2019bertscore}}, a language generation evaluation metric based on pretrained contextual embeddings from BERT~\citep{devlin2019bert}. BERTScore computes the similarity
of two sentences as a sum of cosine similarities between their tokens’ embeddings.
    \item{\textbf{Embedding Similarity}}, which computes the similarity between the text embeddings of two instructions. We use RoBERTa-Large~\citep{liu2019roberta} as the embedding model.
    \item{\textbf{Reward Model ($LM_{\phi}$)}}. We investigate whether the same reward model can be used to produce a similarity score. Given the original and inferred instructions, we directly prompt the model to output a similarity score on a scale from 0 to 1.
    \item{\textbf{Larger Language Model}}. We further experiment with a larger general-purpose language model, prompting it to output a similarity score for a given pair of instructions.
\end{itemize}
Results are reported in Table~\ref{tab:similarity_ablation}. Overall, word-level metrics, while being the simplest and most computationally efficient, outperform embedding-based and LM-based approaches. The consistently poor performance of LM-based methods further highlights the limited ability of language models to perform fine-grained judgment. Among all metrics, the F1 score achieves the best performance across all reward models and evaluation subsets, while remaining the most lightweight and simple.

Surprisingly, word-level F1 achieves better performance than embedding-based metrics, which are generally expected to better capture semantic similarity. Based on our manual annotation, we hypothesize that F1 surpasses BERTScore in this task for several reasons:
\begin{itemize}
    \item \textbf{Sensitivity to related but over- or mis-specified instructions.} In many cases, a response is not preferred because it addresses a related but different request, or because it over-extends beyond the intended scope. This often leads to an inferred instruction that is semantically similar but not exactly aligned with the original instruction. In such cases, F1 appropriately penalizes lexical differences, whereas BERTScore may still assign a high score due to embedding similarity. An example could be “List three causes of the French Revolution” and “Give an overview of the French Revolution, focusing on its causes”.
    \item \textbf{Instructions are short.} Instructions and inferred instructions are typically short. In this regime, lexical overlap metrics such as F1 tend to be more stable, while embedding-based metrics may introduce noise and reduce discrimination.
    \item \textbf{Importance of key entities and exact wording.} Instructions often contain critical nouns (e.g., names, locations, dates, or specific objects). If these tokens are missing or altered, the instruction meaning can change substantially. F1 captures these differences directly, while BERTScore may still give a high similarity score if the sentences remain broadly related. For example, “New York City” and “Los Angeles” have high embedding similarity but they are entirely different. 
\end{itemize}
\vspace{-2mm}

\begin{table*}[h!]
\centering
\renewcommand{\arraystretch}{1.1}
\resizebox{0.7\linewidth}{!}{%
\begin{tabular}{lccccc}
\toprule[1.5pt]
\textbf{Similarity Function} &
\textbf{Qwen3-0.6B} & 
\textbf{Qwen3-1.7B} & 
\textbf{Qwen3-4B} & 
\textbf{Qwen3-8B} & 
\textit{\textbf{Average}}\\
\midrule[1.0pt]
\textbf{BLEU}
&34.4&33.3&35.4&36.6&34.9 \\
\textbf{ROUGE-L}
&34.6&33.6&35.4&\underline{37.2}&35.2 \\
\textbf{F1(BLEU, ROUGE-L)}
&35.1&33.4&35.8&37.0&35.3 \\
\textbf{BERTScore}
&\underline{35.8}&\textbf{35.7}&\underline{35.9}&37.1&\underline{36.1} \\
\textbf{Embed(RoBERTa-L)}
&26.6&27.4&28.9&28.0&27.7 \\
\textbf{the Reward Model}
&12.8&15.3&30.0&26.1&21.0 \\
\textbf{Qwen3-8B}
&29.3&27.0&26.9&26.1&27.3 \\
\textbf{Qwen3-32B}
&31.8&30.6&32.5&32.2&31.8 \\
\rowcolor{gray!20}
\textbf{F1}
&\textbf{36.3}&\underline{35.5}&\textbf{38.1}&\textbf{39.7}&\textbf{37.4} \\
\bottomrule[1.5pt]
\end{tabular}
}
\caption{Ablation study on the choice of similarity function $s$. Reported numbers are RewardBench2 accuracy averaged across all subsets. Columns denote the reward models used to infer the instruction, and rows denote the similarity functions, where “the Reward Model” indicates that the reward model in the corresponding column is also prompted to compute the similarity score. The results show that F1 achieves the best performance while remaining lightweight and simple.}
\label{tab:similarity_ablation}
\vspace{6mm}
\end{table*}

\section{Generator Size}
\label{sec:generator_size}

\begin{figure}[h!]
    \centering
    \includegraphics[width=\linewidth]{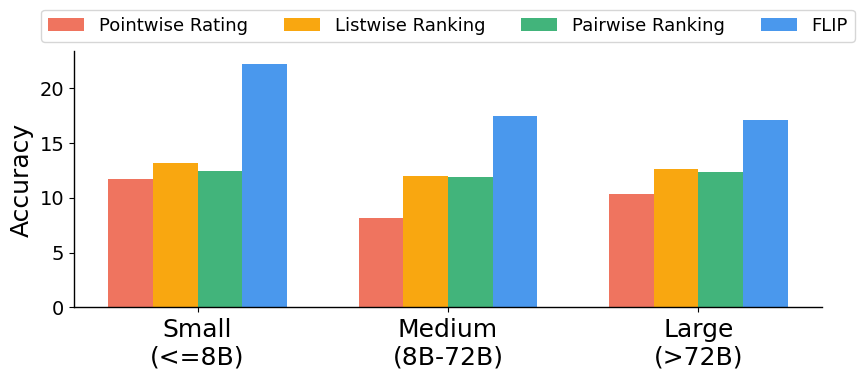}
    \caption{RewardBench2 results across different generator sizes. We only consider instances where all four candidate responses are of the same type.}
    \label{fig:generator_size}
\end{figure}

We conduct a similar analysis with respect to generator model size by selecting instances in which all four candidate responses are produced by generators of similar scale—small, medium, or large in Figure~\ref{fig:generator_size}. 
A noticeable gap to valid responses (i.e., above randomness) remains under this particularly challenging setting, which reflects the increased difficulty of distinguishing fine-grained quality differences among responses generated by similarly capable models. Nevertheless, our method maintains a clear and consistent advantage across all generator scales, underscoring its robustness to generator homogeneity and its effectiveness in difficult evaluation regimes.

\section{Statistical Robustness}
\label{sec:statistical_robustness}
We report accuracy and standard deviations across five runs in Table~\ref{tab:main_sd}. On RewardBench2, \ourmethod{} exhibits slightly higher standard deviations than LLM-as-a-Judge in some cases, which we attribute to the greater expressivity of the generative inference process. Nevertheless, \ourmethod{} consistently outperforms all baselines by a substantial margin across all models, and the performance gaps far exceed the observed run-to-run variability, confirming that the improvements are statistically robust.

\begin{table*}[h!]
    \centering
    \large
    \resizebox{1.0\textwidth}{!}{%
    \renewcommand{\arraystretch}{1}
    \begin{tabular}{llcc|ccc|cccc|ccc|c}
    \toprule[1.5pt]
    \multirow{2}{*}{\textbf{Subset}} &
    \multirow{2}{*}{\textbf{Method}} &
    \multicolumn{2}{c}{\textsc{\textbf{olmo2}}} &
    \multicolumn{3}{c}{\textsc{\textbf{llama3}}} &
    \multicolumn{4}{c}{\textsc{\textbf{qwen3}}} &
    \multicolumn{3}{c}{\textsc{\textbf{gemma3}}} &
    \multicolumn{1}{c}{\textsc{\textbf{mistral}}}\\
    & &
    1B  & 7B 
    & 1B & 3B & 8B
    & 0.6B & 1.7B & 4B & 8B
    & 270M & 1B & 4B
    & 7B  \\
    \midrule[0.75pt]
    
    \multirow{4}{*}{\textbf{\textit{Average}}} 
    & \textbf{Pointwise Rating}
    &13.1&8.9&\underline{16.5}&13.9&15.5&\underline{16.2}&24.0&\underline{29.9}&\underline{32.2}&\underline{16.5}&13.3&17.7&7.9\\
    & (standard deviation)
    &1.1&1.8&1.9&1.8&1.5&2.1&2.3&1.9&1.9&1.5&1.7&2.0&1.2\\[2pt]
    &\cellcolor{gray!20}\textbf{\ourmethod{}} 
    &\cellcolor{gray!20}\textbf{33.1}&\cellcolor{gray!20}\textbf{35.3}&\cellcolor{gray!20}\textbf{29.8}&\cellcolor{gray!20}\textbf{28.2}&\cellcolor{gray!20}\textbf{35.1}&\cellcolor{gray!20}\textbf{36.3}&\cellcolor{gray!20}\textbf{35.5}&\cellcolor{gray!20}\textbf{38.1}&\cellcolor{gray!20}\textbf{39.7}&\cellcolor{gray!20}\textbf{31.2}&\cellcolor{gray!20}\textbf{33.1}&\cellcolor{gray!20}\textbf{35.7}&\cellcolor{gray!20}\textbf{37.3}\\
    & \cellcolor{gray!20}(standard deviation)
    &\cellcolor{gray!20}2.0&\cellcolor{gray!20}2.1&\cellcolor{gray!20}3.5&\cellcolor{gray!20}2.0&\cellcolor{gray!20}2.2&\cellcolor{gray!20}2.4&\cellcolor{gray!20}1.9&\cellcolor{gray!20}1.9&\cellcolor{gray!20}2.3&\cellcolor{gray!20}1.9&\cellcolor{gray!20}2.3&\cellcolor{gray!20}1.6&\cellcolor{gray!20}1.8\\
    \bottomrule[1.5pt]
    \end{tabular}
    }
    \caption{Accuracy and standard deviations of \ourmethod{} and LLM-as-a-Judge baselines using small general-purpose language models on RewardBench2, averaged over five runs. While \ourmethod{} exhibits slightly higher standard deviations than LLM-as-a-Judge in some cases, it consistently outperforms all baselines by a substantial margin across all models, even accounting for run-to-run variability.}
    \label{tab:main_sd}
\vspace{6mm}
\end{table*}

\section{Scaling Up}
\label{sec:scaling_up}
We report the performance of \ourmethod{} and judgment-based baselines on larger general-purpose models in Table~\ref{tab:scalingup}. Although \ourmethod{} becomes more effective as model scale increases, larger models also exhibit stronger judgment capabilities. The relative advantage of employing explicit judgment mechanisms versus \ourmethod{} may depend on the specific model, domain, or task under consideration.

\begin{table*}[h!]
    \centering
    \small
    \resizebox{0.7\textwidth}{!}{%
    \renewcommand{\arraystretch}{1}
    \begin{tabular}{llcc|cc|cc|c}
    \toprule[1.5pt]

    \multirow{2}{*}{\textbf{Subset}} &\multirow{2}{*}{\textbf{Method}} &\multicolumn{2}{c}{\textsc{\textbf{olmo2}}} &\multicolumn{2}{c}{\textsc{\textbf{qwen3}}} &\multicolumn{2}{c}{\textsc{\textbf{gemma3}}} &\multirow{2}{*}{\textbf{\textit{Average}}}\\
    & & 13B  & 32B & 14B & 32B & 12B & 27B& \\
    \midrule[0.75pt]
    
    \multirow{4}{*}{\textbf{Focus}} & \textbf{Pointwise Rating}&7.0&10.3&47.2&40.5&29.6&42.4&29.5\\
    & \textbf{\textit{Listwise Ranking*}}&\underline{24.6}&\underline{30.5}&53.3&\underline{56.2}&\underline{55.1}&\underline{63.8}&47.3\\
    & \textbf{\textit{Pairwise Ranking*}}&20.0&29.0&\underline{59.2}&\underline{56.2}&49.3&56.7&45.1\\
    &\cellcolor{gray!20} \textbf{\ourmethod{}}&\cellcolor{gray!20}\textbf{71.3}&\cellcolor{gray!20}\textbf{67.9}&\cellcolor{gray!20}\textbf{73.0}&\cellcolor{gray!20}\textbf{76.3}&\cellcolor{gray!20}\textbf{65.2}&\cellcolor{gray!20}\textbf{68.1}&\cellcolor{gray!20}\textbf{70.3}\\
    \midrule[0.75pt]

    \multirow{4}{*}{\textbf{Factuality}} & \textbf{Pointwise Rating}&7.5&14.4&\textbf{44.3}&\textbf{37.0}&18.5&23.0&24.1\\
    & \textbf{\textit{Listwise Ranking*}}&23.9&\underline{31.4}&7.6&13.6&\textbf{35.2}&\textbf{39.7}&25.2\\
    & \textbf{\textit{Pairwise Ranking*}}&\underline{27.5}&\textbf{33.3}&\underline{32.8}&\underline{31.4}&\underline{34.0}&\underline{38.6}&\textbf{32.9}\\
    & \cellcolor{gray!20}\textbf{\ourmethod{}}&\cellcolor{gray!20}\textbf{27.7}&\cellcolor{gray!20}25.6&\cellcolor{gray!20}31.1&\cellcolor{gray!20}29.0&\cellcolor{gray!20}24.5&\cellcolor{gray!20}29.3&\cellcolor{gray!20}\underline{27.9}\\
    \midrule[0.75pt]

    \multirow{4}{*}{\textbf{Precise IF}} & \textbf{Pointwise Rating}&4.6&7.9&14.9&\underline{15.1}&5.6&9.0&9.5\\
    & \textbf{\textit{Listwise Ranking*}}&\underline{23.8}&\textbf{25.9}&8.1&10.0&\underline{24.1}&\textbf{29.5}&20.2\\
    & \textbf{\textit{Pairwise Ranking*}}&\textbf{24.4}&\underline{25.8}&\underline{15.8}&13.5&23.6&25.9&\underline{21.5}\\
    & \cellcolor{gray!20}\textbf{\ourmethod{}}&\cellcolor{gray!20}22.2&\cellcolor{gray!20}21.4&\cellcolor{gray!20}\textbf{26.8}&\cellcolor{gray!20}\textbf{26.0}&\cellcolor{gray!20}\textbf{24.4}&\cellcolor{gray!20}\underline{26.2}&\cellcolor{gray!20}\textbf{24.5}\\
    \midrule[0.75pt]

    \multirow{4}{*}{\textbf{Math}} & \textbf{Pointwise Rating}&10.7&22.4&\textbf{40.8}&\textbf{45.0}&26.7&36.6&\underline{30.4}\\
    & \textbf{\textit{Listwise Ranking*}}&\textbf{26.7}&\underline{34.1}&4.2&7.3&\underline{43.4}&\underline{49.3}&27.5\\
    & \textbf{\textit{Pairwise Ranking*}}&25.4&\textbf{41.2}&19.5&22.7&\textbf{48.4}&\textbf{59.2}&\textbf{36.1}\\
    & \cellcolor{gray!20}\textbf{\ourmethod{}}&\cellcolor{gray!20}\underline{26.1}&\cellcolor{gray!20}26.3&\cellcolor{gray!20}\underline{29.6}&\cellcolor{gray!20}\underline{27.2}&\cellcolor{gray!20}22.7&\cellcolor{gray!20}25.7&\cellcolor{gray!20}26.3\\
    \midrule[0.75pt]

    \multirow{4}{*}{\textbf{\textit{Average}}} & \textbf{Pointwise Rating}&7.4&13.8&20.1&27.8&\underline{36.8}&\underline{34.4}&23.4\\
    & \textbf{\textit{Listwise Ranking*}}&\underline{24.8}&30.5&\textbf{39.5}&\textbf{45.6}&18.3&21.8&30.1\\
    & \textbf{\textit{Pairwise Ranking*}}&24.3&\underline{32.3}&\underline{38.8}&\underline{45.1}&31.8&30.9&\underline{33.9}\\
    &\cellcolor{gray!20}\textbf{\ourmethod{}}&\cellcolor{gray!20}\textbf{36.8}&\cellcolor{gray!20}\textbf{35.3}&\cellcolor{gray!20}34.2&\cellcolor{gray!20}37.3&\cellcolor{gray!20}\textbf{40.2}&\cellcolor{gray!20}\textbf{39.6}&\cellcolor{gray!20}\textbf{37.2}\\
    \bottomrule[1.5pt]
    \end{tabular}
    }
    \caption{RewardBench2 accuracy of \ourmethod{} and LLM-as-a-Judge using medium general-purpose language models. The random baseline is 25. Best results are shown in \textbf{bold}, and second-best results are \underline{underlined}. Listwise Ranking has the advantage of observing all candidate completions, Pairwise Ranking observes pairs of completions, while \ourmethod{} and Pointwise Rating operate under the single-completion setting.}
    \label{tab:scalingup}
    \vspace{6mm}
\end{table*}

\section{Test-Time Scaling with Parallel Sampling}
\label{sec:bon-appendix}
\begin{figure*}[h!]
    \includegraphics[width=\linewidth]{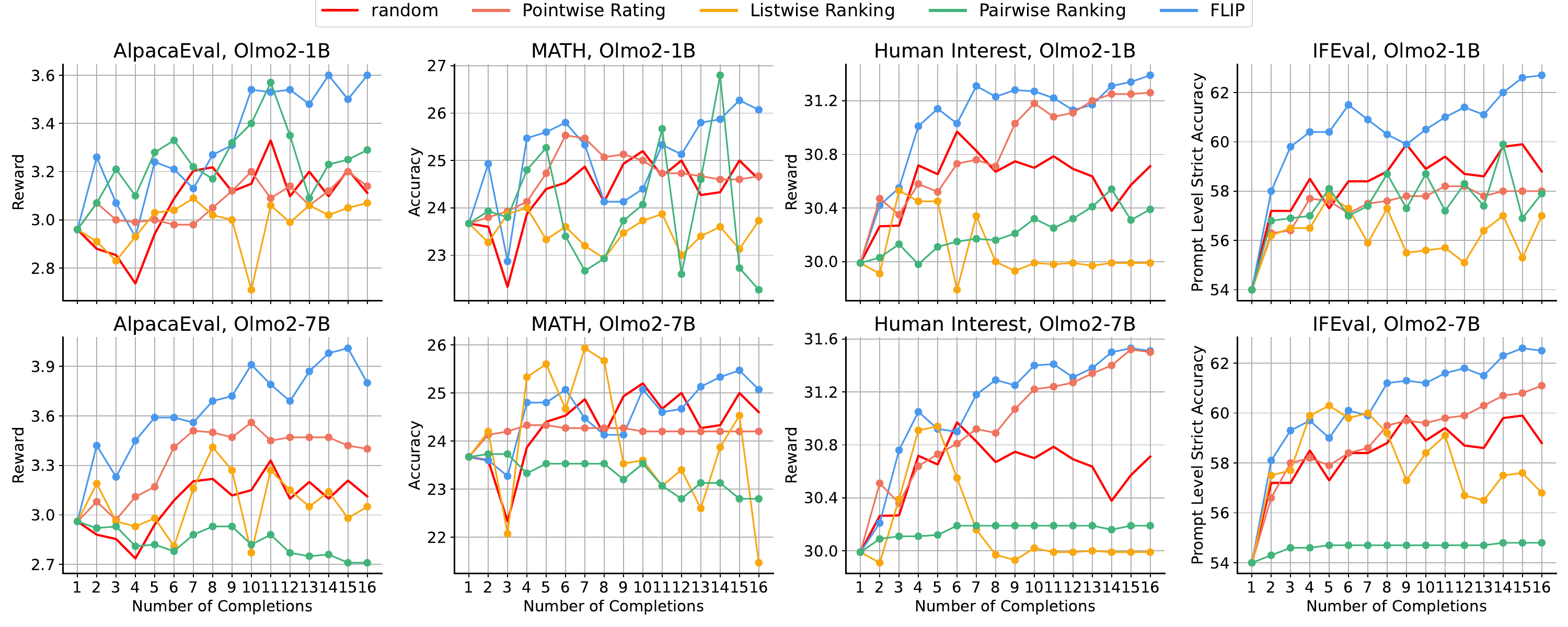}
    \caption{Test-time scaling with parallel sampling on four benchmarks across OLMo 2 families. \ourmethod{} substantially outperforms the baselines, achieving higher performance with greater stability.}
    \vspace{6mm}
    \label{fig:bon-olmo}
\end{figure*}

\begin{figure*}[h!]
    \includegraphics[width=\linewidth]{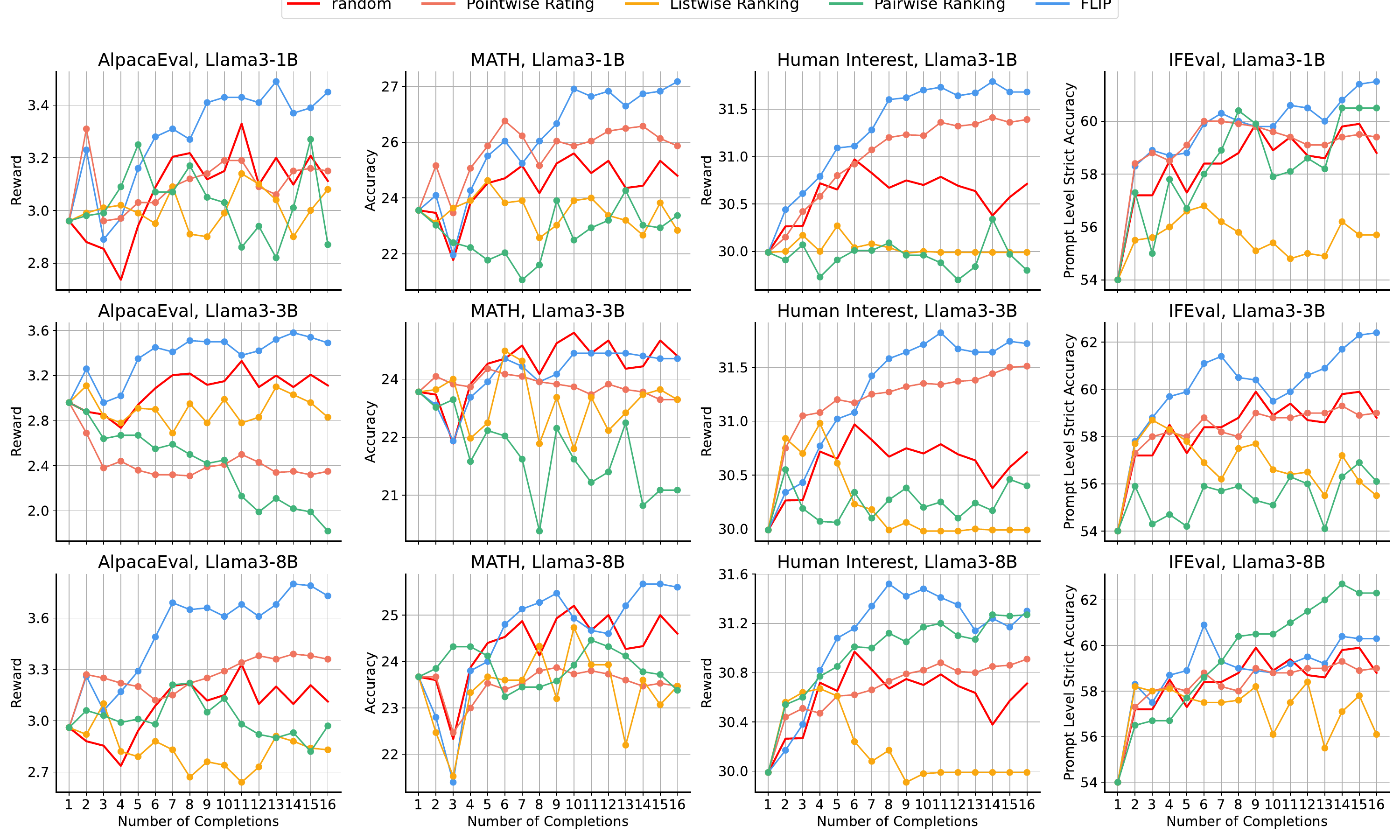}
    \caption{Test-time scaling with parallel sampling on four benchmarks across Llama 3 families. \ourmethod{} substantially outperforms the baselines, achieving higher performance with greater stability.}
    \vspace{6mm}
    \label{fig:bon-llama}
\end{figure*}

Figures~\ref{fig:bon-olmo} and \ref{fig:bon-llama} present detailed results of test-time scaling with parallel sampling using the OLMo 2 and Llama 3 model families as reward models, respectively.

\section{Experiment Details}
\label{sec:experiment_details}

\paragraph{F1 Scores}
When computing F1 scores, both the prediction and ground truth are normalized before scoring: lowercased, stripped of punctuation and articles (a, an, the), and whitespace collapsed. The two normalized strings are then split into words and compared using multiset intersection, so repeated words are counted correctly. Precision measures how many predicted words appear in the ground truth, recall measures the reverse, and F1 is their harmonic mean. If the two word sets share nothing in common, all three metrics return 0.

\paragraph{Intrinsic Evaluation}
Inference experiments are performed on either A100 or H200 GPUs. 
If a response does not follow the instructed output format or results in a tie, it will be considered incorrect and receive a score of 0.

\paragraph{Test-Time Scaling with Parallel Sampling}
In the test-time scaling parallel sampling experiments, we sample completions from Tulu-3-8B-sft instead of Tulu-3-8B for IFEval, as Tulu-3-8B is too high-performing for this experimental setup, following \citet{malik2025rewardbench}.

\paragraph{Reinforcement Learning}
We present hardware configuration in Table~\ref{tab:hardware_configuration_grpo}, batch configuration in Table~\ref{tab:batch_configuration_grpo}, optimization hyperparameters in Table~\ref{tab:optimization_hyperparameters_grpo}, generation parameters in Table~\ref{tab:generation_parameters_grpo}, as well as evaluation details in Table~\ref{tab:evaluation_settings}.

\begin{table*}[h!]
\centering
\resizebox{0.6\textwidth}{!}{%
\begin{tabular}{lc}
\toprule[1.5pt]
Training & 3 nodes × 8 H100 GPUs = 24 GPUs \\
DeepSpeed & ZeRO Stage 3 with gradient checkpointing \\
vLLM Engines & 16 \\
Tensor Parallelism & 1 (single GPU per engine) \\
Learners per node & 8 \\
\bottomrule[1.5pt]
\end{tabular}
}
\caption{Hardware Configuration for the GRPO training.}
\label{tab:hardware_configuration_grpo}
\end{table*}

\begin{table}[h!]
\centering
\begin{tabular}{lc}
\toprule[1.5pt]
Parameter & Value \\
\midrule[1.0pt]
Unique prompts per step & 24 \\
Samples per prompt & 8 \\
Effective batch size & 192 \\
Per-device batch size & 1 \\
Gradient accumulation steps & 8 \\
\bottomrule[1.5pt]
\end{tabular}
\caption{Batch Configuration for the GRPO training.}
\label{tab:batch_configuration_grpo}
\vspace{6mm}
\end{table}

\begin{table}[h!]
\centering
\begin{tabular}{lc}
\toprule[1.5pt]
Parameter & Value \\
\midrule[1.0pt]
Learning rate & $1 \times 10^{-6}$ \\
LR scheduler & Constant (no decay) \\
Warmup steps & 0 \\
Optimizer & AdamW \\
KL coefficient ($\beta$) & 0.0 \\
PPO clip range ($\epsilon$) & 0.272\\
KL estimator & k2 \\
\bottomrule[1.5pt]
\end{tabular}
\caption{Optimization hyperparameters for the GRPO training.}
\label{tab:optimization_hyperparameters_grpo}
\end{table}

\begin{table}[h!]
\centering
\begin{tabular}{lc}
\toprule[1.5pt]
Parameter & Value \\
\midrule[1.0pt]
Response length (max) & 16,384 tokens \\
Prompt length (max) & 2,048 tokens \\
Pack length & 18,432 tokens \\
Temperature (policy) & 1.0 \\
\bottomrule[1.5pt]
\end{tabular}
\caption{Generation parameters for the GRPO training.}
\label{tab:generation_parameters_grpo}
\end{table}

\begin{table*}[h!]
\centering
\resizebox{0.85\textwidth}{!}{%
\begin{tabular}{lccc}
\toprule[1.5pt]
\textbf{Benchmark} & \textbf{Prompting Method} & \textbf{Temperature} & \textbf{Max Token Limit} \\
\midrule[1.0pt]
BBH~\citep{suzgun2023challenging} & Zero-shot & 0.7 & 131,072 \\
GPQA~\citep{rein2024gpqa} & Zero-shot & 0.7 & 131,072 \\
IFEval~\citep{zhou2023instruction} & - & 0.7 & 131,072 \\
IFBench~\citep{pyatkin2025generalizing} & - & 0 & 32,768 \\
Minerva Math~\citep{lewkowycz2022solving} & - & 0.7 & 131,072 \\
\bottomrule[1.5pt]
\end{tabular}
}
\caption{Evaluation settings for GRPO training.}
\label{tab:evaluation_settings}
\end{table*}

\section{Comparison with a Lexical-Overlap Reward Baseline}
\label{app:f1xy_baseline}
 
A question is whether the gains of FLIP stem from backward inference
itself, or merely from the lexical similarity metric: a simple baseline is to
compute F1 (or ROUGE) directly between the instruction and the response,
i.e., $r(x, y) = \mathrm{F1}(x, y)$, without any inference step. We evaluate
this baseline in the Best-of-$N$ setting of Section~\ref{sec:bon_setting}, following
the same experimental protocol. Results across all four benchmarks are
reported in Table~\ref{tab:f1xy_baseline}.
 
\begin{table*}[t]
\centering
\small
\setlength{\tabcolsep}{4pt}
\resizebox{\textwidth}{!}{%
\begin{tabular}{l cccccccccccccccc}
\toprule[1.5pt]
\multirow{2}{*}{\textbf{Method}} & \multicolumn{16}{c}{\textbf{N}} \\
 & 1 & 2 & 3 & 4 & 5 & 6 & 7 & 8 & 9 & 10 & 11 & 12 & 13 & 14 & 15 & 16 \\
\midrule[1pt]
\multicolumn{17}{l}{\textbf{AlpacaEval} (reward)} \\
\midrule[0.75pt]
FLIP
 & \textbf{2.96} & 3.42 & \textbf{3.23} & \textbf{3.45} & 3.59 & \textbf{3.59} & \textbf{3.56} & \textbf{3.69} & \textbf{3.72} & \textbf{3.91} & \textbf{3.79} & \textbf{3.69} & \textbf{3.87} & \textbf{3.98} & \textbf{4.01} & \textbf{3.80} \\
$\mathrm{F1}(x, y)$
 & \textbf{2.96} & \textbf{3.51} & 2.96 & 3.23 & \textbf{3.79} & \textbf{3.59} & 3.48 & 3.39 & 3.35 & 3.29 & 3.41 & 3.33 & 3.33 & 3.35 & 3.29 & 3.15 \\
\midrule[1pt]
\multicolumn{17}{l}{\textbf{MATH} (accuracy)} \\
\midrule[0.75pt]
FLIP
 & \textbf{23.67} & \textbf{24.07} & \textbf{22.47} & \textbf{24.20} & \textbf{25.13} & \textbf{25.53} & \textbf{24.93} & \textbf{25.53} & \textbf{26.00} & \textbf{26.93} & \textbf{26.73} & \textbf{26.87} & \textbf{26.47} & \textbf{26.80} & \textbf{26.87} & \textbf{27.13} \\
$\mathrm{F1}(x, y)$
 & \textbf{23.67} & 23.33 & 20.00 & 21.67 & 22.33 & 24.00 & 23.67 & 23.33 & 22.67 & 24.67 & 25.33 & 24.67 & 25.33 & 25.33 & 24.33 & 22.33 \\
\midrule[1pt]
\multicolumn{17}{l}{\textbf{Human Interest} (reward)} \\
\midrule[0.75pt]
FLIP
 & \textbf{29.99} & \textbf{30.21} & \textbf{30.76} & \textbf{31.05} & \textbf{30.92} & \textbf{30.90} & \textbf{31.18} & \textbf{31.29} & \textbf{31.25} & \textbf{31.40} & \textbf{31.41} & \textbf{31.31} & \textbf{31.38} & \textbf{31.50} & \textbf{31.53} & \textbf{31.51} \\
$\mathrm{F1}(x, y)$
 & \textbf{29.99} & 29.43 & 29.19 & 28.98 & 28.33 & 28.30 & 27.89 & 27.65 & 27.13 & 26.94 & 26.49 & 25.82 & 25.59 & 25.40 & 25.11 & 24.71 \\
\midrule[1pt]
\multicolumn{17}{l}{\textbf{IFEval} (prompt-level strict accuracy)} \\
\midrule[0.75pt]
FLIP
 & \textbf{54.00} & 58.10 & \textbf{59.30} & \textbf{59.70} & \textbf{59.00} & \textbf{60.10} & \textbf{59.90} & \textbf{61.20} & \textbf{61.30} & \textbf{61.20} & \textbf{61.60} & \textbf{61.80} & \textbf{61.50} & \textbf{62.30} & \textbf{62.60} & \textbf{62.50} \\
$\mathrm{F1}(x, y)$
 & \textbf{54.00} & \textbf{58.67} & 57.33 & 56.67 & 55.67 & 55.67 & 56.67 & 56.67 & 56.33 & 54.67 & 53.67 & 53.33 & 52.00 & 53.67 & 52.00 & 52.33 \\
\bottomrule[1.5pt]
\end{tabular}%
}
\caption{Best-of-$N$ results on AlpacaEval, MATH, Human Interest, and
IFEval comparing FLIP against the lexical-overlap baseline
$r(x, y) = \mathrm{F1}(x, y)$. Best results per column within each
benchmark are shown in \textbf{bold}.}
\label{tab:f1xy_baseline}
\end{table*}
 
$\mathrm{F1}(x, y)$ consistently underperforms FLIP across benchmarks,
especially on Human Interest and IFEval, where its performance degrades as
$N$ increases. This confirms that the backward inference step, not merely
the lexical metric, carries the discriminative signal.
 
Beyond the empirical gap, $\mathrm{F1}(x, y)$ is fundamentally unsuitable
as a reward signal. A response that simply copies the instruction verbatim
achieves a perfect score, so the metric provides no information about
response quality and is trivially hackable: optimizing it would directly
drive the policy to echo the instruction rather than answer it. FLIP does
not share this failure mode, as the reward model infers the instruction
from the substantive content of the response.

\section{Format Failures and Valid-Response-Only Accuracy}
\label{app:valid_only}

In our RewardBench2 results (Table~\ref{tab:main}), an output from which
no verdict can be parsed is counted as an error (Appendix~\ref{sec:experiment_details}). A question is how much of FLIP's advantage reflects genuine improvement in judgment quality
rather than robustness to the required output format. We first explain why we
adopt this convention, and then report accuracy computed on valid responses
only.

\paragraph{In deployment, a format failure is a failure.}
A reward model is invoked programmatically inside larger pipelines such as
reranking, parallel sampling, and RL training. When the judge produces an
output from which no signal can be parsed, the pipeline receives no usable
reward at all. A practitioner deploying a small model as a judge experiences
this exactly as an error, so evaluating only on valid responses would
overstate the utility these methods deliver in practice.

\paragraph{Much of the invalidity is itself judgment failure.}
Inspecting the invalid outputs, we find that in many cases the model fails to
produce a parseable verdict precisely because it cannot form a clear
judgment: it declares a tie, hedges between candidates without committing, or
states that it lacks the knowledge required to judge the response. These are
judgment failures expressed through the output format, and we consider it
more faithful to count them as failures than to silently drop them.

That said, separating the two effects is informative. Table~\ref{tab:valid_only}
reports the valid response rate of each method, as well as accuracy computed
on valid responses only, broken down by subset and by model family. All
results are averaged over five independent runs and aggregated within each
model family.

\begin{table}[t]
\centering
\small
\setlength{\tabcolsep}{4pt}
\begin{tabular}{lcccc}
\toprule[1.5pt]
 & \makecell{\textbf{Pointwise}\\\textbf{Rating}} & \makecell{\textbf{\textit{Listwise}}\\\textbf{\textit{Ranking}}} & \makecell{\textbf{\textit{Pairwise}}\\\textbf{\textit{Ranking}}} & \textbf{FLIP} \\
\midrule
\multicolumn{5}{l}{\textit{Valid response rate}} \\
\midrule
OLMo2   & 62.7 & 76.8 & 51.6 & 62.3 \\
Llama3  & 86.1 & 74.6 & 60.8 & 71.5 \\
Qwen3   & 73.7 & 43.5 & 54.7 & 79.7 \\
Gemma3  & 81.9 & 82.7 & 78.6 & 80.7 \\
Mistral & 97.7 & 99.2 & 98.7 & 95.6 \\
\cmidrule(lr){1-5}
Average & 78.6 & 69.1 & 64.5 & 76.6 \\
\midrule[1.0pt]
\multicolumn{5}{l}{\textit{Valid-only accuracy, by subset}} \\
\midrule
Focus      & 26.6 & 40.8 & 35.5 & 61.3 \\
Factuality & 23.3 & 29.5 & 29.7 & 30.7 \\
Precise IF & 19.5 & 29.5 & 28.9 & 29.0 \\
Math       & 37.0 & 37.6 & 37.4 & 39.8 \\
\cmidrule(lr){1-5}
Average    & 26.6 & 34.3 & 32.9 & 40.2 \\
\midrule[1.0pt]
\multicolumn{5}{l}{\textit{Valid-only accuracy, by model family}} \\
\midrule
OLMo2   & 33.0 & 26.3 & 25.3 & 38.3 \\
Llama3  & 17.8 & 28.5 & 26.6 & 34.6 \\
Qwen3   & 39.8 & 50.0 & 47.3 & 49.8 \\
Gemma3  & 19.5 & 25.9 & 26.7 & 34.8 \\
Mistral & 8.5  & 30.8 & 27.8 & 38.4 \\
\cmidrule(lr){1-5}
Average & 26.6 & 34.3 & 32.9 & 40.2 \\
\bottomrule[1.5pt]
\end{tabular}
\caption{Valid response rate (\%) and RewardBench2 accuracy computed on valid
responses only, averaged over five runs and aggregated within each subset or model
family.}
\label{tab:valid_only}
\end{table}

Even when restricted to valid responses, FLIP outperforms Pointwise Rating
(40.2 vs.\ 26.6), Listwise Ranking (40.2 vs.\ 34.3), and Pairwise Ranking
(40.2 vs.\ 32.9) on average. Pairwise Ranking requires multiple rounds of
comparison over all response pairs, and a valid outcome requires every
comparison to parse and the aggregated preferences to yield a unique winner,
which explains its lower valid rate. On \textit{Precise IF} the margin
between FLIP and the ranking methods is small. However, the ranking methods
also drop more instances than FLIP (66.8\% vs.\ 76.6\% valid rate on
average), so their valid-only accuracy is computed on a smaller and likely
easier subset, and they additionally benefit from observing all candidates
jointly, whereas FLIP scores each completion independently. Notably, FLIP's
own valid rate (76.6\%) is comparable to that of Pointwise Rating (78.6\%),
so FLIP is not benefiting from being an easier format to parse. Overall, the
improvement reflects genuine gains in reward quality rather than differences
in format robustness.

Finally, the extrinsic evaluations are immune to this confounder altogether.
In the test-time scaling experiments (Section~\ref{sec:bon_results},
Figure~\ref{fig:bon}) and the GRPO experiments (Section~\ref{sec:grpo_results},
Table~\ref{tab:grpo}), performance is measured directly by downstream
benchmark quality, with no validity-based accounting involved. FLIP
consistently outperforms the judgment baselines in both settings, which
independently confirms that its advantage is not driven by the format failure
convention.

\section{Instruction Echoing in FLIP-Trained Policies}
\label{app:echoing}
 
Because FLIP rewards the similarity between the original and the inferred
instruction, one might worry that optimizing against it incentivizes the
policy to restate the instruction inside its response. We test this directly
by measuring the rate of instruction echoing on IFBench responses, comparing
the base policy (OLMo-3-7B-Think-DPO) against the FLIP-trained policy
(GRPO, step 400). We distinguish three categories: reproducing the full
prompt verbatim, reproducing part of the prompt verbatim, and paraphrasing
the prompt. Results are reported in Table~\ref{tab:echoing}.
 
\begin{table}[t]
\centering
\small
\begin{tabular}{lcc}
\toprule[1.5pt]
Category & Before & FLIP GRPO \\
\midrule
Full verbatim prompt & 5.3  & 2.3 \\
Partial verbatim     & 29.3 & 20.3 \\
Paraphrase           & 34.3 & 35.3 \\
\bottomrule[1.5pt]
\end{tabular}
\caption{Rate (\%) of instruction echoing in IFBench responses, before and
after GRPO training with FLIP as the reward. Echoing does not increase
after training, and verbatim reproduction decreases substantially.}
\label{tab:echoing}
\end{table}
 
Echoing and paraphrasing do not increase after training with FLIP, and
verbatim reproduction in fact decreases: full verbatim echoing drops from
5.3\% to 2.3\% and partial verbatim echoing from 29.3\% to 20.3\%, while the
paraphrase rate is essentially unchanged.
 
We attribute this to the fact that verbatim overlap between the response $y$
and the instruction $x$ carries no first-order payoff under FLIP. What is
rewarded is whatever enables the inferrer to reconstruct $x$ accurately. From
the inferrer's perspective, a restated prompt embedded in a response is
simply additional response text to summarize, and, as discussed in the
Limitations section, the inferrer does not merely parrot embedded instruction
text: the inferred instruction remains dependent on the substantive content
of the response. Moreover, repeating or paraphrasing the instruction often
degrades overall response quality, which further discourages the behavior.

\section{Comparison with Trained Small Reward Models}
\label{app:trained_rm}
 
Our baselines are prompting-based methods that use general-purpose language
models, since FLIP targets the setting where no task-specific reward model is
available. For practitioners, however, it is useful to situate these numbers
relative to off-the-shelf trained reward models at comparable scale. The
strongest publicly available small sequence-classifier RMs are the
Skywork-Reward-V2 series~\citep{liu2024skywork}, which are Bradley--Terry
classifiers. Table~\ref{tab:trained_rm} reports their RewardBench2 accuracy
on the four subsets used in our evaluation, taken from the leaderboard.
 
\begin{table*}[t]
\centering
\small
\setlength{\tabcolsep}{4pt}
\begin{tabular}{llcccc}
\toprule[1.5pt]
Trained RM (BT classifier) & Focus & Factuality & Precise IF & Math \\
\midrule
Skywork-Reward-V2-Qwen3-0.6B  & 79.5 & 58.0 & 40.0 & 71.6 \\
Skywork-Reward-V2-Llama-3.2-1B & 89.3 & 60.8 & 45.6 & 60.1 \\
Skywork-Reward-V2-Qwen3-1.7B  & 88.5 & 65.7 & 44.4 & 72.7 \\
\bottomrule[1.5pt]
\end{tabular}
\caption{RewardBench2 accuracy of off-the-shelf trained small reward models,
reported on the leaderboard, as a reference point for the prompting-based
methods evaluated in this work. The numbers are not strictly comparable to
ours due to a scoring difference: the leaderboard protocol assigns partial
credit to instances with format failures, whereas our protocol scores them as
zero (Appendix~\ref{app:valid_only}).}
\label{tab:trained_rm}
\end{table*}
 
Trained classifier RMs outperform FLIP on RewardBench2. We view the two
approaches as addressing different situations, and offer the following notes
to practitioners.
\begin{itemize}
    \item \textbf{If a well-matched trained RM exists for the target domain, use it.}
Skywork-Reward-V2 is trained on 26 million preference pairs curated through a
large-scale human--LLM annotation pipeline~\citep{liu2024skywork}. Where that
investment has already been made and the domain matches, a trained classifier
is the right tool.
    \item \textbf{If there is no preference data, no annotation budget, or the domains are arbitrary or shifting, use FLIP.}
FLIP requires neither preference data nor references or rubrics, and
consistently outperforms prompting-based judgment baselines in our
experiments.
    \item \textbf{The two are composable rather than competing.}
FLIP requires only a generative model and can be applied on top of any base
model, including one that has itself undergone reward training. Combining
FLIP's backward-inference signal with a trained RM score is a
direction for future work.
\end{itemize}

\section{Ensembling FLIP with Judgment Baselines}
\label{app:ensemble}
 
Since FLIP and LLM-as-a-Judge access different signals, a question is
whether combining them yields a stronger reward model than either alone. For
each of the 13 SLMs, we construct an ensemble reward by taking the majority
vote over Pointwise Rating, Listwise Ranking, Pairwise Ranking, and FLIP, and
re-evaluate on RewardBench2 under the same protocol as Table~\ref{tab:main}
(1,313 instances, four subsets). Results are reported in
Table~\ref{tab:ensemble}.
 
\begin{table*}[t]
\centering
\small
\setlength{\tabcolsep}{4pt}
\resizebox{\textwidth}{!}{%
\begin{tabular}{ll|cc|ccc|cccc|ccc|c|c}
    \toprule[1.5pt]

    \multirow{2}{*}{\textbf{Subset}} &
    \multirow{2}{*}{\textbf{Method}} &
    \multicolumn{2}{c}{\textsc{\textbf{olmo2}}} &
    \multicolumn{3}{c}{\textsc{\textbf{llama3}}} &
    \multicolumn{4}{c}{\textsc{\textbf{qwen3}}} &
    \multicolumn{3}{c}{\textsc{\textbf{gemma3}}} &
    \multicolumn{1}{c}{\textsc{\textbf{mistral}}} &
    \multirow{2}{*}{\textbf{\textit{Average}}}\\
    & &
    1B  & 7B 
    & 1B & 3B & 8B
    & 0.6B & 1.7B & 4B & 8B
    & 270M & 1B & 4B
    & 7B  \\
    \midrule[0.75pt]
\multirow{2}{*}{Focus}
 & FLIP     & \textbf{55.0} & \textbf{65.6} & \textbf{43.3} & \textbf{38.3} & \textbf{64.1} & \textbf{65.7} & \textbf{58.6} & \textbf{66.8} & \textbf{72.2} & \textbf{50.7} & \textbf{59.1} & \textbf{65.5} & \textbf{69.3} & \textbf{59.6} \\
 & Ensemble & 3.7 & 33.8 & 8.4 & 13.4 & 35.1 & 34.1 & 49.1 & 59.3 & 54.2 & 13.9 & 30.5 & 48.6 & 37.8 & 32.5 \\
\midrule
\multirow{2}{*}{Factuality}
 & FLIP     & \textbf{25.8} & \textbf{27.8} & \textbf{26.1} & \textbf{26.4} & \textbf{27.3} & \textbf{28.4} & \textbf{29.6} & \textbf{31.3} & \textbf{31.1} & \textbf{29.8} & \textbf{24.9} & \textbf{26.5} & \textbf{27.1} & \textbf{27.9} \\
 & Ensemble & 3.2 & 18.6 & 9.5 & 15.1 & 21.0 & 16.5 & 23.9 & 23.4 & 24.1 & 11.1 & 16.6 & \textbf{26.5} & 24.3 & 18.0 \\
\midrule
\multirow{2}{*}{Precise IF}
 & FLIP     & \textbf{23.2} & \textbf{21.5} & \textbf{23.0} & \textbf{24.2} & \textbf{23.1} & \textbf{24.6} & \textbf{23.6} & \textbf{25.0} & \textbf{24.4} & \textbf{21.9} & \textbf{22.2} & \textbf{23.8} & 23.0 & \textbf{23.3} \\
 & Ensemble & 3.0 & 15.0 & 9.4 & 12.0 & 15.2 & 12.4 & 15.2 & 10.8 & 12.4 & 9.0 & 19.6 & 17.5 & \textbf{26.0} & 13.7 \\
\midrule
\multirow{2}{*}{Math}
 & FLIP     & \textbf{28.4} & \textbf{26.3} & \textbf{26.6} & \textbf{23.8} & \textbf{25.8} & \textbf{26.3} & \textbf{30.2} & \textbf{29.4} & \textbf{31.0} & \textbf{22.4} & \textbf{26.2} & 27.1 & \textbf{29.7} & \textbf{27.2} \\
 & Ensemble & 2.5 & 12.2 & 8.9 & 16.6 & 25.5 & 10.5 & 29.0 & 18.0 & 15.5 & 5.1 & 21.4 & \textbf{32.0} & 26.9 & 17.2 \\
\midrule
\multirow{2}{*}{Average}
 & FLIP     & \textbf{33.1} & \textbf{35.3} & \textbf{29.8} & \textbf{28.2} & \textbf{35.1} & \textbf{36.3} & \textbf{35.5} & \textbf{38.1} & \textbf{39.7} & \textbf{31.2} & \textbf{33.1} & \textbf{35.7} & \textbf{37.3} & \textbf{34.5} \\
 & Ensemble & 3.1 & 19.9 & 9.1 & 14.3 & 24.2 & 18.4 & 29.3 & 27.9 & 26.6 & 9.8 & 22.0 & 31.2 & 28.8 & 20.4 \\
\bottomrule[1.5pt]
\end{tabular}%
}
\caption{RewardBench2 accuracy of FLIP compared with a majority-vote ensemble
over Pointwise Rating, Listwise Ranking, Pairwise Ranking, and FLIP. The
higher of the two is shown in \textbf{bold} for each model and subset. The
ensemble underperforms FLIP alone in nearly every setting.}
\label{tab:ensemble}
\end{table*}
 
The ensemble reward underperforms FLIP, often by a wide margin: 20.4 versus
34.5 on average, with the gap largest on \textit{Focus} (32.5 versus 59.6).
This reinforces our central claim. For small models, the
validation--generation gap makes judgment unreliable as a reward signal, so
majority voting injects noise rather than complementary information. It also
indicates that FLIP's advantage does not come from capturing a redundant
slice of the judgment signal.
 
Combining FLIP with judgment nonetheless remains an interesting direction,
though it likely requires an \textit{adaptive} rather than a uniform
ensemble. FLIP could be used to catch off-topic, factually incorrect, and
under- or over-addressed responses (Section~\ref{sec:qualitative_examples}), while LM
judges handle dimensions where backward inference is inherently limited, such
as safety-critical queries where the desired behavior is refusal. Ensembling
may also become viable at larger scales, where judgment is competitive
(Appendix~\ref{sec:scaling_up}). Learned or confidence-weighted combinations
of FLIP and judgment for medium and large models are a promising direction
for future work.

\section{Prompts}
\label{sec:prompts}
We list the prompts used in the main sets of experiments in Tables~\ref{tab:prompts1} and~\ref{tab:prompts1_cont}. Additional prompts used for the prompt consistency check are provided in Tables~\ref{tab:prompts_pointwise_rating} through~\ref{tab:prompts_ours}.

\section{Qualitative Examples}
\label{sec:qualitative_examples_appendix}
We present qualitative examples of \ourmethod{} across various RewardBench2 subsets in Table~\ref{tab:example_focus_chosen} to Table~\ref{tab:example_math}. These examples illustrate that our method effectively identifies off-topic, factually incorrect, and instruction-misaligned outputs.

\section{Discussion and Justification}
\subsection{Does FLIP Only Work for Long Responses?}
\label{sec:does_flip_only_work_for_long_responses}
\ourmethod{} relies on the premise that a response contains sufficient length and contextual richness to support backward inference. Importantly, this is precisely the regime in which a reward model is most needed: extremely short responses such as single words or phrases are better handled by simpler metrics such as exact match, and do not require learned reward signals. For the practically relevant setting of open-ended, multi-sentence responses, \ourmethod{} is well suited by design. Empirically, while \ourmethod{} yields the largest gains over LLM-as-a-Judge on long responses, it consistently outperforms all baselines across short and medium responses as well (Figure~\ref{fig:response_length}). This suggests that \ourmethod{}'s advantage is not contingent on response length, but is rather amplified by it.

\subsection{Does FLIP Only Assess Instruction Following?}
A related concern is whether FLIP, by relying on the alignment between the original and inferred instructions, evaluates only instruction-following behavior while neglecting other dimensions of response quality such as factual accuracy, completeness, and depth. We argue that while FLIP is strongest at capturing instruction alignment, it implicitly accounts for these other dimensions as well.

Empirically, the breadth of FLIP's effectiveness is evident across Table~\ref{tab:main}, Figure~\ref{fig:bon} and Table~\ref{tab:grpo}. Under standard benchmarks, test-time scaling, and GRPO training, FLIP consistently outperforms LLM-as-a-Judge not only on instruction-following benchmarks, but also on mathematics (the Math subset of RewardBench2, Minerva Math, and MATH), factuality (the Factuality subset of RewardBench2), reasoning (BBH and GPQA), and open-domain question answering benchmarks (AlpacaEval and Human Interest). This breadth of improvement, spanning domains where instruction alignment is not the primary quality bottleneck, suggests that FLIP captures a richer quality signal than instruction following alone.

Conceptually, as illustrated by the qualitative examples in Section~\ref{sec:qualitative_examples}, FLIP is sensitive to factual accuracy, completeness, and depth through the backward inference mechanism. Factual errors, for instance, often cause the inferred instruction to diverge substantially from the original: a response that identifies "pandas" as the symbolic animal of the United States may lead the model to infer an instruction about China's symbolic animal instead, yielding a low F1 score. Similarly, an incomplete response tends to reconstruct only a subset of the original instruction's requirements, while a shallow response may reconstruct a simpler or underspecified version. Additional examples are provided in Appendix~\ref{sec:qualitative_examples_appendix}.

That said, we acknowledge that instruction recoverability is a proxy for response quality rather than a complete measure of it, in the same way that BLEU measures n-gram overlap as a proxy for translation quality. No proxy is perfect, and both have known failure modes. The value of such proxies lies not in their theoretical completeness, but in their empirical reliability and practical accessibility, particularly in regimes where stronger signals, such as human judgment, verified answers, or large model judges, are unavailable or prohibitively expensive. Our results demonstrate that instruction recoverability is a surprisingly effective proxy across diverse domains, even in cases where its theoretical justification is less direct.

\subsection{Does FLIP Introduce a Verbosity Bias?}
A concern is whether FLIP systematically favors longer responses, since richer context may facilitate more accurate instruction reconstruction. We argue that this is not the case.
FLIP rewards the degree to which a response uniquely and accurately implies its generating instruction. While response length can aid reconstruction, it only does so when the additional content is on-topic. A verbose but off-topic response will reconstruct a different instruction, penalizing the F1 score. Conversely, a concise but precise response can reconstruct the original instruction well. Verbosity therefore helps FLIP only when it reflects genuine quality, which is the desired behavior of a reward model.

Empirically, we find no evidence of verbosity bias in our GRPO experiments. Despite FLIP yielding consistent downstream gains across all benchmarks (Table~\ref{tab:grpo}), we do not observe a corresponding inflation in response length. Since all evaluation benchmarks have verifiable answers, longer responses do not mechanically improve accuracy, so the observed gains must reflect genuine improvements in response quality rather than a length-gaming strategy.

\subsection{What if the Response and Instruction Are in Different Languages?}
In rare cases, a response may be in a different language than the original instruction, for instance, when a model responds in its default language regardless of the instruction's language. This can cause the inferred instruction to be expressed in a different language than the original, rendering word-level F1 an inappropriate similarity metric. In such scenarios, we recommend replacing F1 with an LLM-based similarity judge, which operates on semantic rather than lexical overlap and is therefore robust to cross-lingual mismatches (Appendix~\ref{sec:similarity_function_appendix}). Notably, this issue is limited to cases of genuine language mismatch and does not affect the vast majority of evaluation settings considered in this work.

\subsection{Related Work}
\label{sec:appendix_related_work}
\citet{yerram2024time} introduce Time Reversed Language Models (TRLMs), which score and generate queries conditioned on responses, operating in the reverse direction of time. Specifically, they pre-train a PaLM2 model from scratch in reverse token order and demonstrate that it can complement forward model predictions when used to score the query given a response for re-ranking multiple forward generations. One variant, \textit{TRLM-Fo}, uses the log-probability of the original query given the response as the reranking signal and is most relevant to FLIP. One of the following work LEDOM~\citep{yin2025ledom} pursues a similar idea, introducing \textit{Reverse Reward}, which uses reverse token-level likelihoods from a fully reverse-trained model, combined with the forward LM score, as the reward signal.

However, FLIP differs from both in several key respects. Motivationally, our work explores reward modeling for small language models in low-capacity regimes, whereas these two works are primarily concerned with reverse training itself. Methodologically, FLIP is training-free: it uses general-purpose language models off the shelf, while \textit{Reverse Reward} relies on expensive reverse pre-training from scratch. Furthermore, FLIP uses word-level F1 between the original instruction and the inferred instruction as the reward signal, which operates in a fully black-box manner, whereas \textit{TRLM-Fo} and \textit{Reverse Reward} rely on log-probability signals that require access to model internals. Empirically, \textit{TRLM-Fo} is evaluated on open-ended QA reranking, citation and retrieval tasks, and safety filtering, while we compare against well-established LLM-as-a-Judge baselines across reward benchmarks and test-time scaling settings, and further investigate whether FLIP can serve as a training signal for reinforcement learning. Therefore, even if the core backward-inference idea overlaps, the problem settings, model scales, methodologies, and practical motivations are substantially different.

\begin{table*}[h!]
\begin{tcolorbox}[title=Prompt 1: Pointwise Rating]
{\bf System Prompt}

\#\#\# Task Description

Please act as an impartial judge and evaluate the quality of the answer provided by an AI assistant to the user query displayed below. Judge whether the provided answer is good.

\bigskip
Notes:

- Your evaluation should consider factors such as the helpfulness, relevance, accuracy, depth, creativity, and appropriate level of detail of the response.

- Begin your evaluation by providing a short explanation.

- Be as objective as possible. After providing your short explanation, please output a score on a scale of 1 to 10.

- Please respond in the following JSON format: \{\{"REASONING": "[...]", "SCORE": "<your-score>"\}\}

\tcblower

{\bf User Prompt} 

[Query]

\{prompt\}

\bigskip
[AI Answer]

\{response\}

\bigskip
[Your judgement]

Respond in JSON format. \{\{"REASONING": "[...]", "SCORE": "<your-score>"\}\}

\end{tcolorbox}

\begin{tcolorbox}[title=Prompt 1: Listwise Ranking]
{\bf System Prompt}

\#\#\# Task Description

Please act as an impartial judge and evaluate the quality of the answers provided by AI assistants to the user query displayed below.

\bigskip
Notes:

- Your evaluation should consider factors such as the helpfulness, relevance, accuracy, depth, creativity, and appropriate level of detail of the response.

- Begin your evaluation by briefly explaining your reasoning for selecting the best response.

- Be as objective as possible. After providing your short explanation, output the index (starting from 0) of the best response.

- Please respond in the following JSON format:
\{\{
  "REASONING": "[...your explanation of why the chosen response is best...]",
  "BEST\_RESPONSE\_INDEX": <index-of-best-response>
\}\}

\tcblower

{\bf User Prompt} 

[Query]

\{prompt\}

\bigskip
[AI Answers]

\{response\_list\}

\bigskip
[Your judgement]

Respond in JSON format. \{\{"REASONING": "[...your explanation of why the chosen response is best...]", "BEST\_RESPONSE\_INDEX": <index-of-best-response>\}\}

\end{tcolorbox}

\vspace{-4mm}
\caption{Prompts used for all experiments, unless otherwise stated.}
\label{tab:prompts1}
\end{table*}

\begin{table*}[h!]
\begin{tcolorbox}[title=Prompt 1: Pairwise Ranking]
{\bf System Prompt}

\#\#\# Task Description

Please act as an impartial judge and evaluate the quality of the answers provided by AI assistants to the user query displayed below.

\bigskip
Notes:

- Your evaluation should consider factors such as the helpfulness, relevance, accuracy, depth, creativity, and appropriate level of detail of the response.

- Begin your evaluation by briefly explaining your reasoning for selecting the best response.

- Be as objective as possible. After providing your short explanation, output either "A" or "B" depending on which response is better.

- Please respond in the following JSON format:
\{\{
  "REASONING": "[...your explanation of why the chosen response is best...]",
  "BETTER\_RESPONSE": "A" or "B"
\}\}

\tcblower

{\bf User Prompt} 

[Query]

\{prompt\}

\bigskip
[AI Response A]

\{response1\}

\bigskip
[AI Response B]

\{response2\}

\bigskip
[Your judgement]

Respond in JSON format. \{\{"REASONING": "[...your explanation of why the chosen response is best...]", "BETTER\_RESPONSE":"A" or "B"\}\}

\end{tcolorbox}

\begin{tcolorbox}[title=Prompt 1: \ourmethod{}]
{\bf System Prompt}

\#\#\# Task Description

Please act as an instruction reconstructor. You will receive a model‑generated response. Read it carefully and infer the instruction or prompt that produced it.

\bigskip
Notes:

- Reconstruct a single instruction that would most plausibly generate the given response.

- Specify the user’s intent, required output format, and any constraints.

- Be concise and concrete.

- Do not add information or assumptions beyond what appears in the response.

- Do not overfit to the specific phrasing or examples used in this response.

- Begin with a brief explanation of your reasoning.

- Please respond in the following JSON format: \{\{"REASONING": "<explanation>", "INFERRED INSTRUCTION": "<instruction>"\}\}

\tcblower

{\bf User Prompt} 

[Response]

\{response\}

\bigskip
[Your Inferred Instruction]

Respond in JSON format. \{\{"REASONING": "<explanation>", "INFERRED INSTRUCTION": "<instruction>"\}\}

\end{tcolorbox}

\vspace{-4mm}
\caption{Prompts used for all experiments, unless otherwise stated.}
\label{tab:prompts1_cont}
\end{table*}

\begin{table*}[h!]
\begin{tcolorbox}[title=Prompt 2: Pointwise Rating]
{\bf System Prompt}

\#\#\# Task Description

Please act as an impartial evaluator and assess the quality of the AI assistant’s response to the user query shown below. Determine whether the response is effective.

\bigskip
Notes:

- Your evaluation should take into account factors such as helpfulness, relevance, accuracy, depth, creativity, and whether the level of detail is appropriate.

- Start your evaluation with a brief explanation.

- Remain as objective as possible. After the explanation, assign a score on a scale from 1 to 10.

- Please respond in the following JSON format: \{\{"REASONING": "[...]", "SCORE": "<your-score>"\}\}
\end{tcolorbox}

\begin{tcolorbox}[title=Prompt 3: Pointwise Rating]
{\bf System Prompt}

\#\#\# Task Description

Please serve as an unbiased reviewer and evaluate the quality of the AI assistant’s answer to the user query provided below. Decide whether the response meets an acceptable standard.

\bigskip
Notes:

- Consider criteria such as usefulness, relevance, correctness, thoroughness, creativity, and the appropriateness of the level of detail.

- Begin with a brief justification of your assessment.

- Maintain objectivity throughout the evaluation. After the justification, assign a numerical score from 1 to 10.

- Please respond in the following JSON format: \{\{"REASONING": "[...]", "SCORE": "<your-score>"\}\}
\end{tcolorbox}

\caption{Pointwise Rating prompt variations used in robustness experiments. The user prompt remains unchanged.}
\label{tab:prompts_pointwise_rating}
\end{table*}

\begin{table*}[h!]
\begin{tcolorbox}[title=Prompt 2: Listwise Ranking]
{\bf System Prompt}

\#\#\# Task Description

You are asked to serve as a neutral evaluator and judge the quality of multiple AI-generated responses to the user prompt provided below.

\bigskip
Notes:

- Base your judgment on criteria such as usefulness, relevance to the question, factual correctness, thoroughness, originality, and suitability of detail.

- Begin with a concise justification explaining why one response stands out as superior.

- Maintain objectivity throughout the assessment. After your justification, indicate which response is the best by providing its zero-based index.

- Please respond in the following JSON format:
\{\{
  "REASONING": "[...your explanation of why the chosen response is best...]",
  "BEST\_RESPONSE\_INDEX": <index-of-best-response>
\}\}
\end{tcolorbox}

\begin{tcolorbox}[title=Prompt 3: Listwise Ranking]
{\bf System Prompt}

\#\#\# Task Description

Assume the role of an independent reviewer responsible for comparing several AI assistant replies to the same user question.

\bigskip
Instructions:

- Evaluate each reply using criteria such as practical value, alignment with the prompt, correctness, level of insight, originality, and completeness.

- Open with a short rationale explaining why one reply outperforms the others.

- Keep your judgment fair and unbiased. After the rationale, specify the winning reply by providing its position number, starting from 0.

- Please respond in the following JSON format:
\{\{
  "REASONING": "[...your explanation of why the chosen response is best...]",
  "BEST\_RESPONSE\_INDEX": <index-of-best-response>
\}\}
\end{tcolorbox}

\caption{Listwise Ranking prompt variations used in robustness experiments. The user prompt remains unchanged.}
\label{tab:prompts_listwise_ranking}
\end{table*}

\begin{table*}[h!]
\begin{tcolorbox}[title=Prompt 2: Pairwise Ranking]
{\bf System Prompt}

\#\#\# Task Description

Assume the role of an unbiased reviewer and determine which of the two AI-generated answers best addresses the user’s question.

\bigskip
Criteria to consider:

- Usefulness and relevance to the question

- Factual accuracy

- Completeness and depth of explanation

- Originality or insight where appropriate

- Suitability of the response length and detail

\bigskip
Process:

- Begin with a concise justification explaining why one response outperforms the other.

- Remain neutral and evidence-based in your judgment.

- Conclude by selecting either "A" or "B" as the superior answer.

- Please respond in the following JSON format:
\{\{
  "REASONING": "[...your explanation of why the chosen response is best...]",
  "BETTER\_RESPONSE": "A" or "B"
\}\}
\end{tcolorbox}

\begin{tcolorbox}[title=Prompt 3: Pairwise Ranking]
{\bf System Prompt}

\#\#\# Task Description

You are asked to objectively compare two AI assistant replies to a given user prompt and decide which one is superior.

\bigskip
Assessment considerations:

- How well each response addresses the user’s intent

- Accuracy and reliability of the information

- Clarity, structure, and depth of the explanation

- Added value, insight, or creativity when relevant

- Whether the amount of detail is appropriate

\bigskip
Instructions:

- First, provide a short rationale explaining why the better response stands out.

- Keep the evaluation fair, neutral, and concise.

- Then clearly indicate your choice by selecting "A" or "B".

- Please respond in the following JSON format:
\{\{
  "REASONING": "[...your explanation of why the chosen response is best...]",
  "BETTER\_RESPONSE": "A" or "B"
\}\}
\end{tcolorbox}

\caption{Pairwise Ranking prompt variations used in robustness experiments. The user prompt remains unchanged.}
\label{tab:prompts_pairwise_ranking}
\end{table*}

\begin{table*}[h!]
\begin{tcolorbox}[title=Prompt 2: \ourmethod{}]
{\bf System Prompt}

\#\#\# Task Description

Act as an instruction reverse-engineer. You will be given a response generated by a model. Carefully analyze the response and infer the most likely instruction or prompt that led to it.

\bigskip
Notes:

- Reconstruct a single, plausible instruction.

- Clearly capture the user’s intent, expected output format, and constraints.

- Keep the instruction concise and specific.

- Do not introduce assumptions or details not supported by the response.

- Start with a short explanation of how you derived the instruction.

- Please respond in the following JSON format: \{\{"REASONING": "<explanation>", "INFERRED INSTRUCTION": "<instruction>"\}\}
\end{tcolorbox}

\begin{tcolorbox}[title=Prompt 3: \ourmethod{}]
{\bf System Prompt}

\#\#\# Task Description

Assume the role of an instruction reverse engineer. You will receive a model-generated response and must analyze it to deduce the most likely prompt that produced it.

\bigskip
Notes:

- Infer one coherent and plausible instruction.

- Accurately reflect the user’s intent, required output structure, and any constraints.

- Be concise and precise.

- Avoid adding assumptions or information not evidenced by the response.

- Do not tailor the instruction too closely to the exact examples in the response.

- Begin with a brief explanation describing how the instruction was inferred.

- Please respond in the following JSON format: \{\{"REASONING": "<explanation>", "INFERRED INSTRUCTION": "<instruction>"\}\}
\end{tcolorbox}

\caption{\ourmethod{} prompt variations used in robustness experiments. The user prompt remains unchanged.}
\label{tab:prompts_ours}
\end{table*}

\begin{table*}[h!]
    \centering
    \small
    \begin{adjustbox}{max width=1\textwidth}
    \begin{tabular}{p{1.3in} p{5in}}
         \toprule[1.5pt]
         {\textbf{Instruction}} & \textbf{Reasons for and against the use of the atomic bomb in 1945.} \\
         \midrule[1pt]
         {\textbf{Subset}} & \textbf{Focus} \\
        \midrule[1pt]

         \textbf{Chosen Response} &
         The decision to use atomic bombs on Hiroshima and Nagasaki in August 1945 was a complex one, with both compelling reasons for and against its use. Here are some of the key arguments from each side:
        \bigskip
        
        \#\#\# Reasons for Using the Atomic Bomb
        
        1. **End the War Quickly:**
           - The primary reason cited by U.S. leaders was to end the war quickly and save American lives. They believed that a full-scale invasion of Japan would result in significant casualties on both sides.
        
        2. **Deterrence:**
           - The United States wanted to demonstrate the power of its new weapon to the Soviet Union and other potential adversaries. This could serve as a deterrent for future conflicts.
        
        3. **Psychological Impact:**
           - The hope was that the devastating effects of the atomic bombs would shock the Japanese government into surrendering without the need for further destruction or invasion.
        
        4. **Strategic Advantage:**
           - The U.S. military believed that the atomic bomb would give them a strategic advantage over the Soviet Union, which was also involved in the Pacific theater.
        
        5. **Prevent Further Suffering:**
           - Some argued that using the bomb would prevent even greater suffering and destruction if the war continued, including the potential for a Japanese suicide attack on American cities.
        \bigskip
        
        \#\#\# Reasons Against Using the Atomic Bomb
        
        1. **Casualties and Moral Concerns:**
           - There were significant concerns about the massive civilian casualties that would result from the bombings. The estimated death toll from the atomic bombs was in the tens of thousands, and many more would suffer long-term health effects.
        
        2. **International Repercussions:**
           - The use of atomic weapons raised serious moral and ethical questions. It set a precedent for the use of such devastating weapons and could have long-lasting negative impacts on international relations.
        
        3. **Alternative Methods:**
           - Some argued that there were alternative methods to force Japan's surrender, such as a demonstration of the bomb's power in an uninhabited area or a formal warning to the Japanese government.
        
        4. **Impact on the Soviet Union:**
           - There were concerns that the use of the bomb might provoke the Soviet Union to accelerate its own nuclear program, leading to an arms race.
        
        5. **Long-Term Consequences:**
           - The long-term environmental and health consequences of radioactive fallout were not fully understood at the time but were later recognized as significant issues.
        
        6. **Japanese Surrender Without Bombing:**
           - Historians debate whether Japan would have surrendered without the atomic bombs. Some argue that the Potsdam Declaration and the Soviet declaration of war on August 8th were sufficient to prompt Japan's surrender.
        \bigskip
        
        \#\#\# Conclusion
        
        The decision to use the atomic bomb remains one of the most controversial in modern history. While it is clear that the U.S. leadership believed it was necessary to end the war quickly and save lives, the moral and ethical implications continue to be debated. The use of these weapons marked a turning point in human history and had profound and lasting effects on global politics and warfare.\\
        \midrule[1pt]
        \textbf{Inferred Instruction} & Write a balanced analysis of the decision to use atomic bombs on Hiroshima and Nagasaki, highlighting both the reasons for and against the use of these weapons, and present the argument in a neutral tone.\\
         \midrule[1pt]
         \textbf{F1(Instruction, Inferred Instruction)} & 0.341 \\

         \bottomrule[1.5pt]
    \end{tabular}
    \end{adjustbox}
    \caption{Example of \ourmethod{} on the RewardBench2 \textit{Focus} subset. See Table~\ref{tab:example_focus_rejected} for the rejected response. The chosen response closely follows the instruction and discusses the reasons for and against the use of the atomic bomb, whereas the rejected response deviates by focusing on public opinion before and after the atomic bomb. \ourmethod{} successfully captures this deviation, yielding higher similarity between the inferred instruction based on the chosen response and the original instruction than that based on the rejected response.}
    \label{tab:example_focus_chosen}
\end{table*}

\begin{table*}[h!]
    \centering
    \small
    \begin{adjustbox}{max width=1\textwidth}
    \begin{tabular}{p{1.3in} p{5in}}
         \toprule[1.5pt]
         {\textbf{Instruction}} & \textbf{Reasons for and against the use of the atomic bomb in 1945.} \\
         \midrule[1pt]
         {\textbf{Subset}} & \textbf{Focus} \\
        \midrule[1pt]

         \textbf{Rejected Response} &
         Public opinion in the United States underwent significant changes before and after the atomic bombings of Hiroshima and Nagasaki in August 1945. Here's a brief overview of the shift in public opinion:
        \bigskip
        
        **Before the bombings (1945):**
        
        1. **Support for the use of atomic bombs:** In the months leading up to the bombings, there was a growing sense of urgency and frustration among Americans regarding the war in the Pacific. Many people believed that the use of atomic bombs would bring a swift end to the war and save American lives. A Gallup poll conducted in June 1945 found that 85\% of Americans supported the use of atomic bombs if it would shorten the war.
        
        2. **Fear of the Japanese:** The Japanese government had been refusing to surrender, and there was a growing fear among Americans that the Japanese would continue to fight to the last man, leading to a prolonged and bloody conflict. This fear contributed to the widespread support for the use of atomic bombs.
        
        3. **Scientific and military enthusiasm:** The scientific community and the military were excited about the potential of the atomic bomb, and many saw it as a game-changer in the war. The Manhattan Project, the secret research and development project that produced the atomic bomb, was shrouded in secrecy, and many Americans were unaware of the devastating consequences of the bomb.
        \bigskip
        
        **After the bombings (August 1945):**
        
        1. **Shock and horror:** The immediate reaction to the bombings was one of shock and horror. The news of the devastating effects of the atomic bombs on Hiroshima and Nagasaki was met with widespread outrage and dismay. Many Americans were unaware of the full extent of the destruction and the number of civilian casualties.
        
        2. **Changing attitudes towards the use of atomic bombs:** As the full extent of the damage became clear, public opinion began to shift. Many Americans began to question the morality and necessity of using such a devastating weapon. A Gallup poll conducted in September 1945 found that 37\% of Americans believed that the use of atomic bombs was a mistake, while 44\% still supported it.
        
        3. **Growing concern about the future of nuclear warfare:** The bombings raised concerns about the potential for future nuclear conflicts and the devastating consequences of such a war. This led to a growing debate about the ethics and consequences of nuclear warfare, which would continue to shape American public opinion in the years to come.
        
        4. **Rise of the anti-nuclear movement:** The bombings marked the beginning of the anti-nuclear movement in the United States. Many Americans began to question the use of nuclear energy and the development of nuclear weapons, leading to a growing movement to ban nuclear testing and reduce the threat of nuclear war.

        \bigskip
        
        In summary, public opinion in the United States shifted from a general support for the use of atomic bombs to a growing concern about the devastating consequences of such a weapon. The bombings marked a turning point in American public opinion, leading to a greater awareness of the risks and consequences of nuclear warfare.
         \\
        \midrule[1pt]

         \textbf{Inferred Instruction}&
         Write a comprehensive overview of the shift in public opinion in the United States before and after the atomic bombings of Hiroshima and Nagasaki in 1945, including the reasons for the shift and the impact on American public opinion. \\
        \midrule[1pt]
         \textbf{F1(Instruction, Inferred Instruction)}  &
         0.318 \\
         \bottomrule[1.5pt]
    \end{tabular}
    \end{adjustbox}
    \caption{Example of \ourmethod{} on the RewardBench2 \textit{Focus} subset (continued).}
    \label{tab:example_focus_rejected}
\end{table*}

\begin{table*}[h!]
    \centering
    \small
    \begin{adjustbox}{max width=1\textwidth}
    \begin{tabular}{p{1.3in} p{5in}}
         \toprule[1.5pt]
         {\textbf{Instruction}} & \textbf{Write a list of 20 animated films from the 1990s, in the following format: “[TITLE] ([YEAR]), by [DIRECTORS]”. Do not write any descriptions, or sort in any order.} \\
         \midrule[1pt]
         {\textbf{Subset}} & \textbf{Factuality} \\
        \midrule[1pt]

         \textbf{Chosen Response} &
         1. The Lion King (1994), by Roger Allers and Rob Minkoff
         
        2. Beauty and the Beast (1991), by Gary Trousdale and Kirk Wise
        
        3. Aladdin (1992), by Ron Clements and John Musker
        
        4. Toy Story (1995), by John Lasseter
        
        5. Pocahontas (1995), by Mike Gabriel and Eric Goldberg
        
        6. The Nightmare Before Christmas (1993), by Henry Selick
        
        7. Mulan (1998), by Tony Bancroft and Barry Cook
        
        8. A Bug's Life (1998), by John Lasseter and Andrew Stanton
        
        9. Hercules (1997), by Ron Clements and John Musker
        
        10. The Hunchback of Notre Dame (1996), by Gary Trousdale and Kirk Wise
        
        11. Anastasia (1997), by Don Bluth and Gary Goldman
        
        12. The Prince of Egypt (1998), by Brenda Chapman, Steve Hickner, and Simon Wells
        
        13. Antz (1998), by Eric Darnell and Tim Johnson
        
        14. Tarzan (1999), by Chris Buck and Kevin Lima
        
        15. The Iron Giant (1999), by Brad Bird
        
        16. James and the Giant Peach (1996), by Henry Selick
        
        17. The Secret of NIMH 2: Timmy to the Rescue (1998), by Dick Sebast
        
        18. Quest for Camelot (1998), by Frederik Du Chau
        
        19. Thumbelina (1994), by Don Bluth and Gary Goldman
        
        20. The Swan Princess (1994), by Richard Rich\\
        \midrule[1pt]
        \textbf{Inferred Instruction} & List 20 popular animated films from the 1990s that were produced by Disney, Pixar, or other animation studios.\\
         \midrule[1pt]
         \textbf{F1(Instruction, Inferred Instruction)} & 0.381 \\
        \midrule[1pt]
         \midrule[1pt]

         \textbf{Rejected Response} &
         "Beauty and the Beast (1991), by Gary Trousdale and Kirk Wise
         
        "The Lion King (1994), by Roger Allers and Rob Minkoff
        
        "Aladdin (1992), by Ron Clements and John Musker
        
        "Toy Story (1995), by John Lasseter
        
        "Matilda (1996), by Danny DeVito
        
        "Anastasia (1997), by Don Bluth and Gary Goldman
        
        "Hercules (1997), by Ron Clements and John Musker
        
        "Mulan (1998), by Tony Bancroft and Barry Cook
        
        "Tarzan (1999), by Chris Buck and Kevin Lima
        
        "Antz (1998), by Eric Darnell and Tim Johnson
        
        "Monsters, Inc. (2001), by Pete Docter
        
        "The Prince of Egypt (1998), by Brenda Chapman, Steve Hickner, and Simon Wells
        
        "Flubber (1997), by Les Mayfield
        
        "Kiki's Delivery Service (Majo no Takkyūbin) (1998), by Hayao Miyazaki
        
        "A Bug's Life (1998), by John Lasseter and Andrew Stanton
        
        "Stuart Little (1999), by Rob Minkoff
        
        "Treasure Planet (2002), by Ron Clements and John Musker
        
        "Titan A.E. (2000), by Don Bluth and Gary Goldman
        
        "James and the Giant Peach (1996), by Henry Selick
        
        "The Hunchback of Notre Dame (1996), by Gary Trousdale and Kirk Wise"
         \\
        \midrule[1pt]

         \textbf{Inferred Instruction}&
         List the 20 most popular animated films released between 1991 and 2005, with a focus on American and American-influenced films. \\
        \midrule[1pt]
         \textbf{F1(Instruction, Inferred Instruction)}  &
         0.186 \\
         \bottomrule[1.5pt]
    \end{tabular}
    \end{adjustbox}
    \caption{Example of \ourmethod{} on the RewardBench2 \textit{Factuality} subset. The chosen response correctly lists 20 films from the 1990s, whereas the rejected response incorrectly includes several films from the 2000s. \ourmethod{} successfully captures this difference, yielding higher similarity between the inferred instruction based on the chosen response and the original instruction than that based on the rejected response.}
    \label{tab:example_factuality}
\end{table*}

\begin{table*}[h!]
    \centering
    \small
    \begin{adjustbox}{max width=1\textwidth}
    \begin{tabular}{p{1.3in} p{5in}}
         \toprule[1.5pt]
         {\textbf{Instruction}} & \textbf{What's the root meaning of politics Please use an emoji at the end of every sentence.} \\
         \midrule[1pt]
         {\textbf{Subset}} & \textbf{Precise IF} \\
        \midrule[1pt]

         \textbf{Chosen Response} &
         The root meaning of politics comes from the Greek word "politikos," which means "of, for, or relating to citizens." [EMOJI] It originally referred to the art or science of government and the activities associated with governing a city-state, like Athens. [EMOJI] Over time, the term has evolved to encompass all aspects of government, including the strategies used to gain or maintain power and the decisions that influence the welfare of a community or nation. [EMOJI]\\
        \midrule[1pt]
        \textbf{Inferred Instruction} & Explain the meaning and development of the term 'politics' over time, including an emoji representation of politics.\\
         \midrule[1pt]
         \textbf{F1(Instruction, Inferred Instruction)} & 0.370 \\
        \midrule[1pt]
        \midrule[1pt]

         \textbf{Rejected Response} &
         The root meaning of "politics" comes from the Greek word "politikos," which means "of, for, or relating to citizens." This word is derived from "polis," which means city or state. The study of politics involves the distribution of power, governance, public policy, and the methods and tactics used to formulate policies. [EMOJI]\\
        \midrule[1pt]

         \textbf{Inferred Instruction}&
         Define politics, including its origins and main topics of study.\\
        \midrule[1pt]
         \textbf{F1(Instruction, Inferred Instruction)}  &
         0.174 \\
         \bottomrule[1.5pt]
    \end{tabular}
    \end{adjustbox}
    \caption{Example of \ourmethod{} on the RewardBench2 \textit{Precise IF} subset. The chosen response correctly adds an emoji at the end of every sentence, whereas the rejected response does not. \ourmethod{} successfully captures this difference, yielding higher similarity between the inferred instruction based on the chosen response and the original instruction than that based on the rejected response.}
    \label{tab:example_precise_if}
\end{table*}

\begin{table*}[h!]
    \centering
    \small
    \begin{adjustbox}{max width=1\textwidth}
    \begin{tabular}{p{1.3in} p{5in}}
         \toprule[1.5pt]
         {\textbf{Instruction}} & \textbf{It would be necessary to convert the mass of $\,\_ \times 10^{9}$ protons into energy in order to release $63.1\ \mathrm{J}$ of energy.} \\
         \midrule[1pt]
         {\textbf{Subset}} & \textbf{Math} \\
        \midrule[1pt]

         \textbf{Chosen Response} &
        Let
        
        1) We'll use Einstein's mass-energy equivalence equation: $E = mc^2$
        
        \quad where $E$ is energy in Joules (J) 
        
        \quad $m$ is mass in kg 
        
        \quad $c$ is speed of light $= 3 \times 10^8$ m/s
        
        2) We know: 
        
        \quad $E = 63.1$ J 
        
        \quad We need to find the number of protons
        
        3) Rearranging $E = mc^2$: 
        
        \quad $m = \dfrac{E}{c^2}$ 
        
        \quad $m = \dfrac{63.1}{(3 \times 10^8)^2}$ 
        
        \quad $m = \dfrac{63.1}{9 \times 10^{16}}$ 
        
        \quad $m = 7.01 \times 10^{-16}$ kg
        
        4) To convert from kg to protons: 
        
        \quad Mass of one proton $= 1.67 \times 10^{-27}$ kg 
        
        \quad Number of protons $= \dfrac{\text{total mass}}{\text{mass of one proton}}$ 
        
        \quad $= \dfrac{7.01 \times 10^{-16}}{1.67 \times 10^{-27}}$ 
        
        \quad $= 4.20 \times 10^{11}$
        
        5) Since the answer needs to be in terms of $x \times 10^9$: 
        
        \quad $4.20 \times 10^{11} = 420 \times 10^9$
        
        Therefore, $420 \times 10^9$ protons would need to be converted to energy to release $63.1$ J of energy.
        
        The answer is 420.\\
        \midrule[1pt]
        \textbf{Inferred Instruction} & Calculate the number of protons (in multiples of $10^{9}$) required to release $63.1$~Joules of energy using the mass--energy equivalence equation and the mass of a proton.\\
         \midrule[1pt]
         \textbf{F1(Instruction, Inferred Instruction)} & 0.419 \\
        \midrule[1pt]
        \midrule[1pt]

         \textbf{Rejected Response} &
         To solve this problem, we need to use the Einstein's mass--energy equivalence formula: $E = mc^2$, where $E$ is the energy, $m$ is the mass, and $c$ is the speed of light.
        
        We are given the energy $E = 63.1~\text{J}$. We need to find the mass $m$ that would release this energy when converted into energy using Einstein's formula.
        
        First, we need to convert the speed of light c into meters per second. The speed of light is approximately $c = 3 \times 10^8~\text{m/s}.$
        
        Now, we can rearrange the formula $E = mc^2$ to solve for mass $m$: $m = \frac{E}{c^2}.$
        
        Plugging the given energy $E = 63.1~\text{J}$ and the speed of light $c = 3 \times 10^8~\text{m/s}$ into the formula, we get: $m
          = \frac{63.1~\text{J}}{9 \times 10^{16}{m^2/s^2}}
          = 7.011 \times 10^{-16}~\text{kg}.$
        
        Now, we know that 1 proton has a mass of approximately $1.67 \times 10^{-27}~\text{kg}.$ To find out how many protons have a mass of $7.011 \times 10^{-16}~\text{kg}$, we can divide the total mass by the mass of one proton:
        
        Number of protons = $\frac{7.011 \times 10^{-16}~\text{kg}}{1.67 \times 10^{-27}~\text{kg/proton}}
          = 4.20 \times 10^{11}~\text{protons}.$
        
        So, it would be necessary to convert the mass of $\boxed{4.20 \times 10^{11}}$ protons into energy in order to release $63.1~\text{J}$ of energy.

        The answer is: $4.20 \times 10^{11}$.
        \\
        \midrule[1pt]

         \textbf{Inferred Instruction}&
         Given an energy value (63.1 J), find out how many protons have a combined mass that, when converted to energy, would result in the given energy using Einstein's mass-energy equivalence formula.\\
        \midrule[1pt]
         \textbf{F1(Instruction, Inferred Instruction)}  &
         0.375 \\
         \bottomrule[1.5pt]
    \end{tabular}
    \end{adjustbox}
    \caption{Example of \ourmethod{} on the RewardBench2 \textit{Math} subset. The chosen response correctly converts the final answer into multiples of $10^9$, whereas the rejected response does not. \ourmethod{} successfully captures this difference, yielding higher similarity between the inferred instruction based on the chosen response and the original instruction than that based on the rejected response.}
    \label{tab:example_math}
\end{table*}

\end{document}